\documentclass[default,iicol]{sn-jnl}

\usepackage{subfig}
\usepackage{bm}
\usepackage{pifont}
\usepackage{makecell}
\usepackage[export]{adjustbox}
\usepackage{bbm}

\jyear{2021}%

\theoremstyle{thmstyleone}%
\theoremstyle{thmstyletwo}%

\theoremstyle{thmstylethree}%

\begin{document}

\title[Article Title]{A Generalized Framework for Predictive Clustering and Optimization}

\author*[1]{\fnm{Aravinth} \sur{Chembu}}\email{aravinth.chembu@mail.utoronto.ca}

\author[1]{\fnm{Scott} \sur{Sanner}}\email{ssanner@mie.utoronto.ca}

\affil[1]{\orgdiv{Department of Mechanical and Industrial Engineering}, \orgname{University of Toronto}, \orgaddress{\street{27 King's College}, \city{Toronto}, \postcode{M5S 1A1}, \state{ON}, \country{Canada}}}



\abstract{Clustering is a powerful and extensively used data science tool.
While clustering is generally thought of as an unsupervised learning technique, there are also supervised variations such as   
Spath's clusterwise regression that attempt to find clusters of data that yield low regression error on a supervised target.
We believe that clusterwise regression is just a single vertex of a largely unexplored design space of supervised clustering models. In this article, we define a generalized optimization framework for predictive clustering that admits different cluster definitions (arbitrary point assignment, closest center, and bounding box) and both regression and classification objectives.  
%
We then present a joint optimization strategy that exploits mixed-integer linear programming (MILP) for global optimization in this generalized framework. To alleviate scalability concerns for large datasets, we also provide highly scalable greedy algorithms inspired by the Majorization-Minimization (MM) framework. Finally, we demonstrate the ability of our models to uncover different interpretable discrete cluster structures in data by experimenting with four real-world datasets.}

\keywords{Supervised clustering, mixed-integer linear programming, Majorization-Minimization, Data science}



\maketitle

\section{Introduction}

The availability of massive volumes of data coupled with the need to understand, analyze and explore patterns in them as a means to find solutions and drive decision making has made clustering a popular tool in data science. Cluster analysis is widely used in problems with unlabeled data and has become synonymous to unsupervised learning. It has been found helpful in a varied range of machine learning and data mining tasks, including pattern recognition, document clustering and retrieval, image segmentation, and medical and social sciences~\cite{Jain_survey, Rui_xu, Xu_Tian, Wang2017,Naik2017}. This reflects its broad applicability, and usefulness as an exploratory data analysis tool, especially for large datasets. However, relatively little focus has been directed towards using clustering for predictive tasks with labeled data.

In many cases, it is natural to assume that real data is generated from complex processes that might be mixtures of discrete modes of a predictive target response.  Some of these modes could just be from different processes that generate the data~\cite{Huang, Brusco_remarks, Vicari, SilvaWclr, silvaCLR}, while others may be due to implicit or explicit confounders that lead to a significant change in the response variable being predicted. 
Naturally, a single predictive model cannot capture such multiple relationships between the dependent and explanatory variables.


For an example use case, consider the housing price regression predictive task for a city. Crime rates influence the housing market, and in most cases, property values drop with an increase in crime~\cite{Ceccato, Boggess}. But in contrast to this trend, housing values in inner-city or downtown areas are high regardless of the high crime rates. This positive relationship could be because of increased reporting and higher property crimes in affluent high-income neighborhoods~\cite{Boggess, Lynch, Ceccetosweden}. Clearly, one regression model cannot capture these two different trends in prices with respect to the crime rate. This kind of multiple regression modeling has been found suitable for analyzing data from various domains, including housing price prediction~\cite{geosciences}, marketing analysis~\cite{desarbo88}, demographic neighborhood analysis~\cite{olson}, and weather prediction~\cite{bagirov}.





\begin{figure*}[t]
\begin{centering}
\subfloat[Three different cluster definitions]{\includegraphics[width=10.7cm]{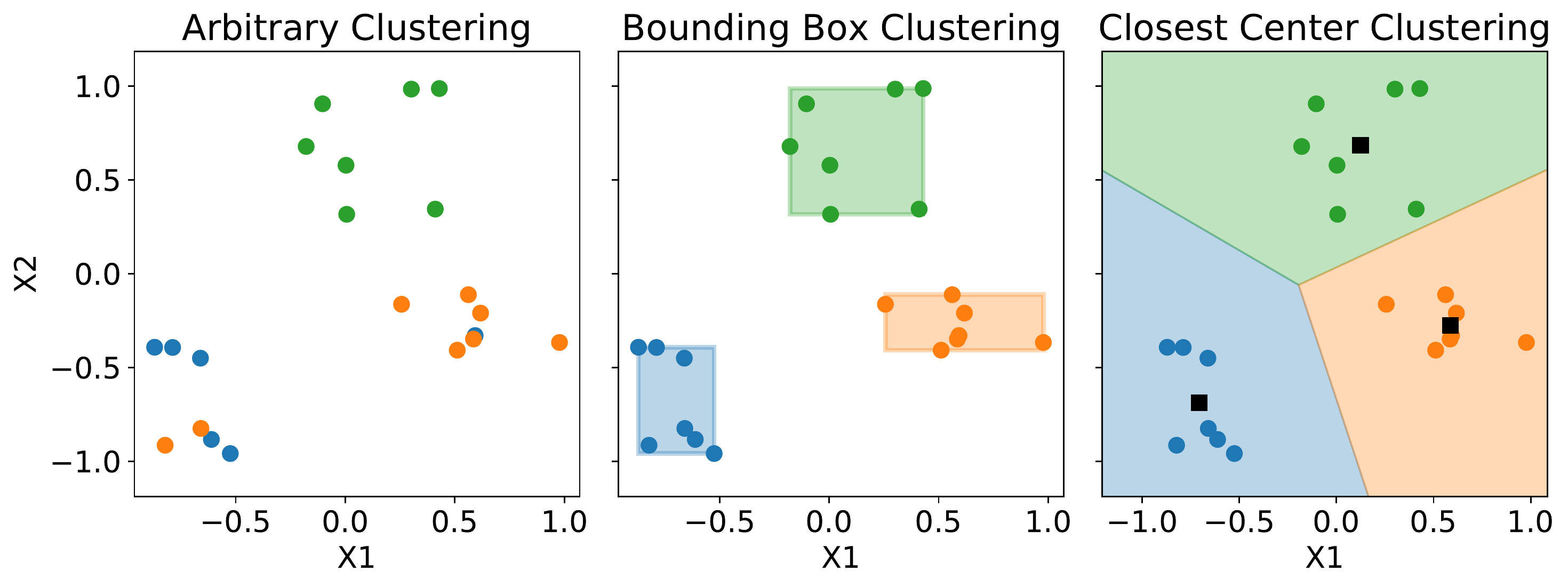}\label{fig:exampleClusters}}\hspace{0.09cm}
\subfloat[Synthetic data]{\includegraphics[width = 4.7cm]{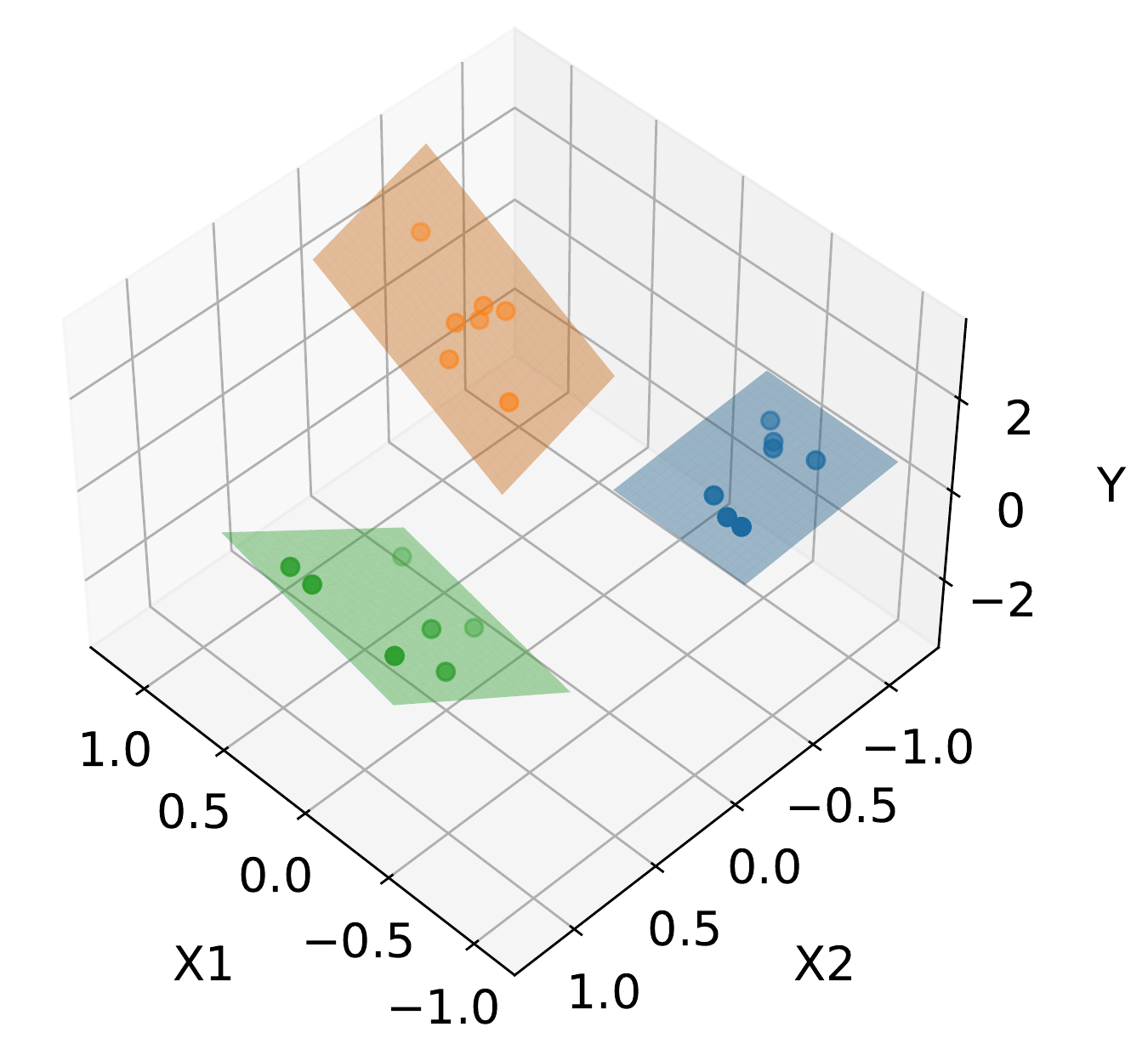}\label{fig:examplePoints}}   
\par\end{centering}
\caption{Illustrative example with multiple generative modes that were uncovered with predictive clustering. (a) Different perspectives to clustering: (1) arbitrary clustering with points assigned without any restriction (left), (2) clusters defined as bounding boxes w.r.t. to features (center), and (3) clusters defined as regions closest to data centers demonstrated here using a Voronoi plot with the three cluster centers represented by points shown in black (right).
(b) Synthetic data with three distinct regression planes (dependent variable Y as a linear function of two independent variables X1, X2) where the projection of points on the X1-X2 feature space gives well-separated clusters as shown by blue, orange, and green points (right).    
}
\label{fig:example}
\end{figure*}

To this end, historically, several methods have gone beyond standard unsupervised clustering to supervised or predictive versions. Most of these models from the literature fall under the clusterwise regression (CLR) category~\cite{Spath79, crio, opc, Manwani}. These models primarily aim to identify disjoint subsets or explicit subclasses of the data that lead to different predictive (regression in this case) models in each cluster. However, existing methods for predictive clustering are largely bespoke for specific problems, supervised objectives, or cluster definitions and have largely gone unused as a general tool for data science. This motivated us to take a broader perspective towards clustering and build a framework to explore the predictive clustering design space.

Consider, for example, the samples of points shown in Figure~\ref{fig:example}. We generated this data consisting of points from three different regression planes such that the points in these three disjoint groups are reasonably well separated in the feature space. Predictive clustering aims to identify these distinct modes present in the data. The plots show multiple perspectives of clustering with the supervised regression objective used to solve this problem. We can either (1) assign data to clusters without any restriction on the search space (as is the case with traditional CLR~\cite{Spath79, crio}), (2) define clusters as bounding boxes in the feature space, or (3) define clusters as the regions nearest to exemplar data centers~\cite{opc, Manwani}. The plots with the projection of points in the feature plane show how these clustering methods differ and identify the three groups. We remark that to date, clustering methods have been defined for (1) and (3), but only limited approximated options are available when it comes to (2)~\cite{breiman1984classification, oct}.

In this article, we seek to comprehensively explore the design space of supervised objectives and cluster definitions that allow us to identify several important gaps in this space. To this end, we formalize a general mathematical framework for predictive clustering that subsumes existing methods and introduces new ones. We also propose global optimization methods that can directly exploit our unifying formalization as well as general greedy optimization methods that are highly scalable for large-scale datasets, near-optimal on cases where we can compare to global methods, and which reduce to existing methodologies in some special cases. Finally, we demonstrate the power of this unified perspective through a variety of applications that exhibit how different supervised objective and cluster definitions allow us to detect and learn important discrete structures and behaviors in the dataset. 

We summarize our main contributions in the article as follows:

\begin{itemize}
    \item We present a general framework for predictive clustering that combines clustering with a supervised objective. Specifically, we focus on three clustering methods as shown in Figure~\ref{fig:exampleClusters}, and we call them arbitrary, closest center, and bounding box clustering. Furthermore, we explore two supervised loss functions in our design space for regression and classification tasks. We identify that clusterwise classification adds novelty in the field of general linear classification.
    \item We provide two ways for optimizing the loss functions in our models: (1) mixed-integer linear programming (MILP) for global optimization, and (2) greedy methods inspired by the Majorization-Minimization (MM) prescription of algorithms to tackle the scalability issues of using MILP, and at the same time provide comparable but sub-optimal (locally optimal) solutions.
    \item We demonstrate the applicability of the different models in our framework with case studies on four real-world datasets and evaluate its performance with baseline linear models. These models provide highly interpretable results that we believe will help in decision and policy making when applied to data science problems.
\end{itemize}

\section{Related work}

There is a substantial body of research related to clustering and its applications in unsupervised learning tasks. However, our proposed contributions focus more on clustering as a predictive tool. Therefore, we briefly survey the literature on available clustering techniques, followed by relevant research focused on using clustering for supervised learning tasks. 


{\bf K-means and alternative cluster definitions:} Clustering is an extensively researched topic in the context of advancements in clustering techniques (engineer highly scalable and fast algorithms) and its applications in problems in data science. Since surveying this sheer mass of literature is beyond the scope of this article (some comprehensive clustering surveys~\cite{Jain_survey, Rui_xu, Xu_Tian}), we focus only on several clustering techniques relevant to our work.


The most commonly used method for cluster analysis, especially in the context of hard-partition clustering, is the popular K-means algorithm~\cite{Lloyd}. It is a fast heuristic algorithm designed to solve the minimum sum-of-squares clustering problem (MSSC), where the task is to choose clusters such that the points within clusters have small sum-squared errors. 
Several attempts have been made to solve the MSSC problem optimally using column generation and integer linear programming~\cite{BruscoRBB, duMerle, Aloise, Tiasguns}; however, none of these could scale like K-means. 



Similarly, several other definitions for clusters exist in the literature. Among them, density-based clustering~\cite{DBSCAN, Optics} has gained huge popularity primarily because of its ability to produce arbitrary-shaped clusters in contrast to k-means which can only deal with spherical clusters. 
Yet another approach is defining clusters based on grids as first described in the CLIQUE~\cite{clique} algorithm for clustering high dimensional data.
Here, the central idea was to first discretize the entire space into a mesh with a predefined grid size followed by identifying grids with a dense collection of points in subspaces. Although our bounding boxes clustering method (refer to Figure~\ref{fig:exampleClusters}) resembles the grid-based definitions for a cluster, they differ in how these clusters are identified. CLIQUE uses a bottom-up approach - unions of dense cells from lower subspaces to define clusters in higher dimensions. In contrast, our model directly identifies the bounding boxes based on a supervised optimization objective.

{\bf Predictive Clustering:} Numerous methods have been mentioned in the literature that have moved away from discussing clustering in the traditional sense and have focused on using it for predictive purposes. However, most of these models were designed for a specific supervised learning objective or application. Therefore, we survey the supervised clustering literature in two directions: clusterwise methods for regression and classification.




{\bf Clusterwise Regression (CLR) greedy models:} The central idea in clusterwise regression (CLR) is to split the data into several disjoint sets to identify the various regression modes present in them. In the pioneering work of Sp\"{a}th~\cite{Spath79} in CLR, he proposed an exchange method to jointly optimize the overall regression error by unifying regression and clustering phases. In this approach, two observations from different clusters would be exchanged if it reduces the overall error. In follow-up work, Sp\"{a}th~\cite{Spath81} proposed a faster exchange algorithm where a single observation is shifted between clusters if it reduces the overall cost. More recently, Manwani and Sastry~\cite{Manwani} proposed the K-plane regression algorithm, which is similar, in spirit, to the K-means~\cite{Lloyd} algorithm. This approach repeatedly involves (1) identifying the best regression weights in each cluster and (2) reassigning each observation to have the least error when assigned to that cluster. The above heuristic approaches provide acceptable solutions in many cases; however, as with K-means, they are sensitive to initializations and converge to sub-optimal solutions.

{\bf Optimal CLR methods:} Several researchers have tried to provide globally optimal solutions for the CLR problem. Lau et al.~\cite{Lau} proposed a nonlinear programming formulation for a variant to the CLR problem, but they do not provide any guarantees for the optimal solution. A more common approach seen in the literature starts from the CLR problem’s quadratic programming (QP) reformulation. As a first, Carbonneau et al.~\cite{Carb} proposed a mixed-logical quadratic programming formulation to solve the CLR problem to global optimality feasibly. They further improved upon this approach in their later works~\cite{Carbb, Carbcg} where they used linear integer programming tricks such as column generation and repetitive branch and bound~\cite{BruscoRBB} methods coupled with heuristic algorithms.


\begin{table*}[t]
\caption{Summary of relevant works from the literature that are either directly consumed in our framework or are approximate solutions to our models}   
\begin{minipage}{\columnwidth}
\begin{center}
\begin{tabular}{llll}
    \toprule
        & \multicolumn{3}{c}{Cluster Assignment} \\	\cmidrule{2-4}
        
    	&	Arbitrary	&	Closest Center	&	Bounding Boxes	\\	\midrule
    MSSC	&	-	&	K-means~\cite{Lloyd} 	&	-	\\	
    Regression	&	CLR~\cite{Spath79}, CRIO~\cite{crio}	&	OPC~\cite{opc}, K-plane~\cite{Manwani} 	&	Model Trees~\cite{Modeltrees}, ORT~\cite{ORT}	\\	
    Classification	&	-	&	-	&	Classification Trees~\cite{breiman1984classification}, OCT~\cite{oct}	\\	\bottomrule
\end{tabular}
\end{center}
\end{minipage}
\label{tab:relMethods}
\end{table*}

As an alternative approach, Bertsimas and Shioda~\cite{crio} proposed the CRIO model where they used the more robust absolute error metric (similar to Sp\"{a}th's model in ~\cite{Spath86}) as the regression loss and used MILP to solve the problem optimally. In more recent research, Zhu et al.~\cite{zhuclr} also adopt the same approach for CLR. Obviously, this approach is more elegant and computationally less intensive than the QP counterparts. Here, we remark that the CLR method, specifically CRIO~\cite{crio}, is captured in our framework under arbitrary clustering (refer to Table~\ref{tab:relMethods}). This approach to clustering works reasonably well for some instances, especially when the regression lines from two different clusters intersect. However, as shown in Figure~\ref{fig:exampleClusters}, arbitrary assignment fails to identify the three well-separated clusters. Thus, homogeneity among points in clusters w.r.t. to the feature variables can be a desirable trait, as argued by several researchers in their works on CLR~\cite{Brusco_remarks, Vicari, silvaCLR, opc, Manwani}.


To address this drawback and obtain homogeneous clusters, Manwani and Sastry~\cite{Manwani} expanded on their work to present a modified K-plane regression algorithm. In this approach, the authors added the MSSC loss (w.r.t. to the independent variables) to the squared error regression loss (with a regularization parameter). The same approach was used by Silva et al.~\cite{silvaCLR}. In more recent research, authors in~\cite{opc} presented the Optimal Predictive Clustering (OPC) method where they used a variant of the cost function used in the above approaches. They included the dependent variable (along with the features) while computing the MSSC error per cluster. Furthermore, they provided a greedy algorithm based on K-means++~\cite{kmeansPlus} to warm-start the mixed-integer quadratic programming method to obtain near-optimal results. These approaches are arguably similar to our regression with the closest center clustering method. However, in contrast to OPC~\cite{opc}, we use the absolute error cost function for regression and hence, obtain a computationally more tractable MILP formulation for global optimization. Moreover, we provide a different methodology for greedy optimization when compared with the modified K-plane regression algorithm~\cite{Manwani}. 

We also present a novel bounding box clustering methodology to solve the CLR problem. With this approach, not only do we retain the critical advantage of the closest center method of having coherent clusters, but we also identify a set of decision rules to define a cluster. This adds to the interpretability of our results. This may be in a similar vein to model trees~\cite{Modeltrees, wangm5}, where a greedy approach is used to build decision trees with regression models at the leaves. Also, to solve model trees optimally, Bertsimas and Dunn~\cite{ORT} presented the Optimal Regression trees with linear predictions (ORT-L) model. Fundamentally, these approaches build a decision tree in search of good regression fits at the leaves (with just binary splits on a single feature at each node); hence, we believe that this approximates our approach. In contrast, our model performs a more holistic search of the feature space to identify the best set of bounding boxes. 

{\bf Clusterwise Classification:} Much of the research at the intersection of clustering and classification has been along the lines of either cluster-and-classify or clustering-based classification. In the former approach, clustering precedes the classification task. One such approach clustered large datasets to a relatively smaller number of clusters and used the cluster centroids to complete the classification task~\cite{Evans}. Other methods were more application-specific, where the data was first clustered, and then a classification model was run on each cluster. Fahad et al.~\cite{Fahad} used this approach for activity recognition in smart homes; Tammenah et al.~\cite{Taamneh} used hierarchical clustering with Neural networks to classify road traffic accidents.


In contrast to the cluster-and-classify approaches, cluster-based classifiers perform the classification task assisted by clustering. Bertsimas and Shoda~\cite{crio}, in their CRIO, assigned one class of points to clusters such that no points of the other class belong in these clusters. The drawback of this approach is that it can only address a binary classification problem. Furthermore, clustering assisted information retrieval and text classification are also common~\cite{cbc_text, cbc_semisuper}. In more recent research, clustering was used for information retrieval to find multiple clusters that hold highly relevant retrieved information~\cite{relevance_driven}. All these approaches identify clusters such that all observations in them belong to the class of interest.

In this work, we focused on a per cluster classification model approach, similar in spirit to the CLR approach. 
Unlike the cluster-and-classify approaches, our clusterwise classification (CLC) models jointly optimize the overall error of the clustering and classification tasks. Moreover, we use the closest center and bounding box clustering for our CLC tasks. 
The bounding box approach for CLC can be seen as similar to classification trees~\cite{breiman1984classification} and their optimal versions called optimal classification trees~\cite{oct}. However, in contrast to classification tree methods which partition the feature space to propose one class per leaf, our approaches have one classification model per partitioned space.

In summary, with our framework, we were able to identify and address critical gaps in the supervised clustering literature while simultaneously capturing some existing models like CLR and CRIO (refer Table~\ref{tab:relMethods}). Overall, in our work: (1) we directly capture the MILP based CRIO approach and K-plane regression greedy algorithm; (2) we provide a different problem formulation and loss function (MAE, which is more robust) for the OPC model; (3) we propose a greedy optimization strategy that is different from the modified K-plane regression algorithm; (4) we describe an alternative approach to CLR with our bounding boxes clustering model and (5) we present a novel clusterwise classification approach.

\section{Methodology}

In this section, we formally present the predictive clustering framework and the mathematical notations we use. We then describe our two optimization procedures: (1) Mixed Integer optimization (MIO) and (2) greedy algorithms.

\subsection{Problem definition}

As the name suggests, the framework for predictive clustering constitutes of two main ``ingredients'': 
\begin{enumerate}
    \item {\bf Prediction:} involves optimizing a supervised objective function to predict the label or the dependent variable. Typical loss functions we include are mean squared error (MSE) and mean absolute error (MAE) for linear regression tasks, and hinge loss for soft margin support vector machines (SVM) for both binary and multi-class classification tasks. 
    \item {\bf Cluster assignment:} which involves assigning every observation in the data to a cluster based on an assignment choice.  As previously mentioned, the different options available are arbitrary (Arbit), closest center (CC), and bounding boxes (BB) clustering (refer to the Figure~\ref{fig:exampleClusters}). 
\end{enumerate}
In the following subsections, we describe the above-mentioned clustering methods and loss functions in detail.

\subsection{Notation}
 
We assume that we have $N$ observations of the form $D = (\bm{x_{i}}$,$y_{i})$ (for $i \in \overline{N} = \{1,...,N\})$ in the data, and where $\bm{x_{i}}$ is the feature variable vector and $y_{i}$ is the label to be predicted. Moreover, we assume that the features $\bm{x_i} \in \mathbb{R}^d$ are $d$ dimensional ($\bm{x_i} = (x_{i1}, ..., x_{id})$). We note that clustering without any prediction is the trivial case when labels $y_{i}$ corresponding to all observations are null. The goal in hard-partitioning clustering is to assign each of the $N$ observations to one of the $K$ clusters $\{C_1, ...,C_K\}$ where $K \leq N$. We also have binary indicator variables $c_{ik}$ to identify cluster assignments for all observations in the data. If a point $i$ is associated with cluster $C_k$, then we have $c_{ik} = 1$; $c_{ik} = 0$, otherwise. With the cluster definitions as above, we desire the following properties:

\begin{itemize}
    \item No overlap between clusters: $C_k \cap C_j = \phi, \;\; \forall \; k,j \; \in \overline{K} \;\; \text{and}  \;\; k \neq j $ 
    \item All observations are assigned to clusters: $\bigcup_{i=1}^K C_i = D$
\end{itemize}

We now define notations to capture various cluster definitions and supervised loss functions. We use the following notation throughout the rest of the article:

\begin{itemize}
    \item Variables $\bm{\theta_k}$ to denote the cluster-specific parameters of our model. It can be the weights of the regression planes or weights defining the hyperplanes separating the classes in a classification task.
    \item Per-datum error represented by $l( \bm{x_{i}}$,$y_{i}, \bm{\theta_k})$ to typically indicate, for instance, the hinge loss in the case of SVM or squared error for regression associated with each observation.
    \item Overall error $L(\bm{\theta,c})$ for any combination of clustering and supervised objective function given by: 
    \begin{equation}
        L(\bm{\theta,c}) = \sum_{k=1}^{K}\sum_{i=1}^{N} l( \bm{x_{i}},y_{i}, \bm{\theta_k})  \; c_{ik}
        \label{eqn:totalloss}
    \end{equation}
    The per-datum error $l( \bm{x_{i}},y_{i}, \bm{\theta_k})$ is multiplied by the indicator variable $c_{ik}$ to ensure that for each observation we only account for the error associated with the cluster it is assigned to.
\end{itemize}
    
\begin{table*}[t]
\caption{Summary of the predictive clustering design space.  We exclude the combination of Arbitrary cluster assignment for Hinge loss (SVM) since having overlapping SVMs increases complexity and hence reduces the interpretability of the model.}   
\begin{minipage}{\textwidth}
\begin{center}
\begin{tabular}{l l l l}
    \toprule
        & \multicolumn{3}{c}{Clustering Type} \\	\cmidrule{2-4}
        
    	&	Arbitrary	&	Closest Center	&	Bounding Box	\\	\midrule
    MAE/MSE regression loss	&	\ding{51}	&	\ding{51} 	&	\ding{51}	\\	
    Hinge loss for SVM	&		&	\ding{51}	&	\ding{51}	\\	\bottomrule
\end{tabular}
\end{center}
\end{minipage}
\label{tab:designspace}
\end{table*}

\subsection{Supervised learning objective}

In this subsection, we discuss the supervised error functions we used in our framework.

\begin{enumerate}
\item{\bf Regression loss :} The central idea in CLR is to cluster the data while simultaneously learning cluster-specific regression models through a join optimization methodology. We can either use mean squared error (MSE) or mean absolute error (MAE) to optimize our regression. With MAE, the total error would be given by:

\begin{equation}
    \min_{c,\bm{\theta} }\; \; \; \sum_{k=1}^{K} \sum_{i=1}^{N} \rvert y_{i} - \bm{\theta_{k}'x_{i}}\rvert c_{ik} 
\end{equation}

Here, $\bm{\theta_{k}}$ stands for the regression weights in the $k$-th cluster and the per-datum loss for this case is $l( \bm{x_{i}},y_{i}, \bm{\theta_k}) = \rvert y_{i} - \bm{\theta_{k}'x_{i}}\rvert $. Similarly, for the MSE loss function, we have $l( \bm{x_{i}},y_{i}, \bm{\theta_k}) = (y_{i} - \bm{\theta_{k}'x_{i}})^2$. While both error measures are quite similar, MAE penalizes the outliers less substantially and hence, is more robust than MSE. Moreover, the overall CLR problem with the MAE loss reduces to a MILP formulation which can be more tractable and computationally less expensive than a Quadratic Programming formulation with MSE loss.

\item{\bf Classification loss :} Similar to CLR, the purpose of CLC is to group points and run per-cluster classification models to drive the overall classification error to a minimum. This article only focuses on multi-class classification with SVM, wherein we find the hyperplanes that best separate the multiple classes found in each cluster.

The classical approach to solve the multi-class problem with SVM is to employ a collection of binary classifiers with the one-vs-all classification trick~\cite{rifkin}. However, for our MILP formulations, we utilize the first ``single machine" approach for the multi-class case called Weston and Watkins (WW-SVM)~\cite{WWSVM}. This approach provides a single error value per data which we could then elegantly plug into our framework. For a $M$ class classification task, the overall cost function with this approach along with an L1 regularization~\cite{l1SVM} for the coefficients is given by: 

\begin{equation}
\begin{split}
    &\min_{c,\bm{\theta} }\; \; \;\sum_{k=1}^{K} \sum_{m=1}^{M} \|\bm{\theta_{k, m}}\|_1 + C \sum_{k=1}^{K} \sum_{i=1}^{N} \sum_{m \neq y_{i}} \xi_{i}^{m}  c_{ik} \\
    &\;\;\text{s.t.} \;\; \;   \bm{\theta_{k,y_{i}}}'\bm{x_{i}} \geq \bm{\theta_{k,m}}'\bm{x_{i}} + 2 - \xi_{i}^{m}, \\
            &\;\; \; \quad \quad m = \{1, ...,M\}\backslash y_i 
    \label{eqn:clsloss1}
\end{split}
\end{equation}


Here, the per-datum loss would be given by $l( \bm{x_{i}},y_{i}, \bm{\theta_k}) = \sum_{m \neq y_{i}} \xi_{i}^{m}$.  


\end{enumerate}

\subsection{Cluster assignment}


In this subsection, we briefly introduce the three unique clustering methods currently included in our framework. We provide the mathematical formulations necessary to achieve these cluster definitions in Section~\ref{section:mio} along with a mixed-integer optimization procedure to solve the overall model.

\begin{enumerate}
    \item {\bf Arbitrary clustering (Arbit):} The assignment of a point to a cluster is independent of any constraints, and optimizing the supervised learning objective drives these assignments. For example, when we combine regression and arbitrary clustering, we obtain the traditional CLR model~\cite{crio, Spath79}. The main advantage with arbitrary clustering is its ability to find overlapping clusters, specifically, intersecting regression lines in the case of CLR. However, as noted previously with the synthetic data example in Figure~\ref{fig:example}, this method fails to identify the three well-separated clusters but instead provides overlapping clusters as its solution. Hence, we seek other clustering methods that provide within-cluster homogeneity.
    \item {\bf Closest center (CC) clustering:} In this clustering method, points are assigned to their closest cluster center to give spherical-shaped coherent clusters in the feature space. The central idea is to determine the best $K$ cluster centers such that assigning observations in the data to them minimizes the overall loss function. The cluster centers help interpret and analyze the profile of points that belong to a cluster.  
    \item {\bf Bounding boxes (BB) clustering:} The fundamental idea is to define clusters as axis-parallel hypercuboids (rectangles in the two features case as shown in Figure~\ref{fig:exampleClusters}), and observations that fall within the boundaries of a cluster belong to that cluster. A key benefit is that clusters can now be characterized with a set of decision rules like a DNF expression, making the models highly interpretable.
\end{enumerate}

\begin{table*}[t!]
\caption{The overall MILP formulations for the three clustering types}   
\begin{minipage}{\textwidth}
\begin{center}
\begin{tabular}{c | l}
    \toprule
        
    	Clustering	&	MILP Formulation	\\	\midrule

    \makecell[c]{Arbitrary  \\	  \includegraphics[scale=0.3, valign=t]{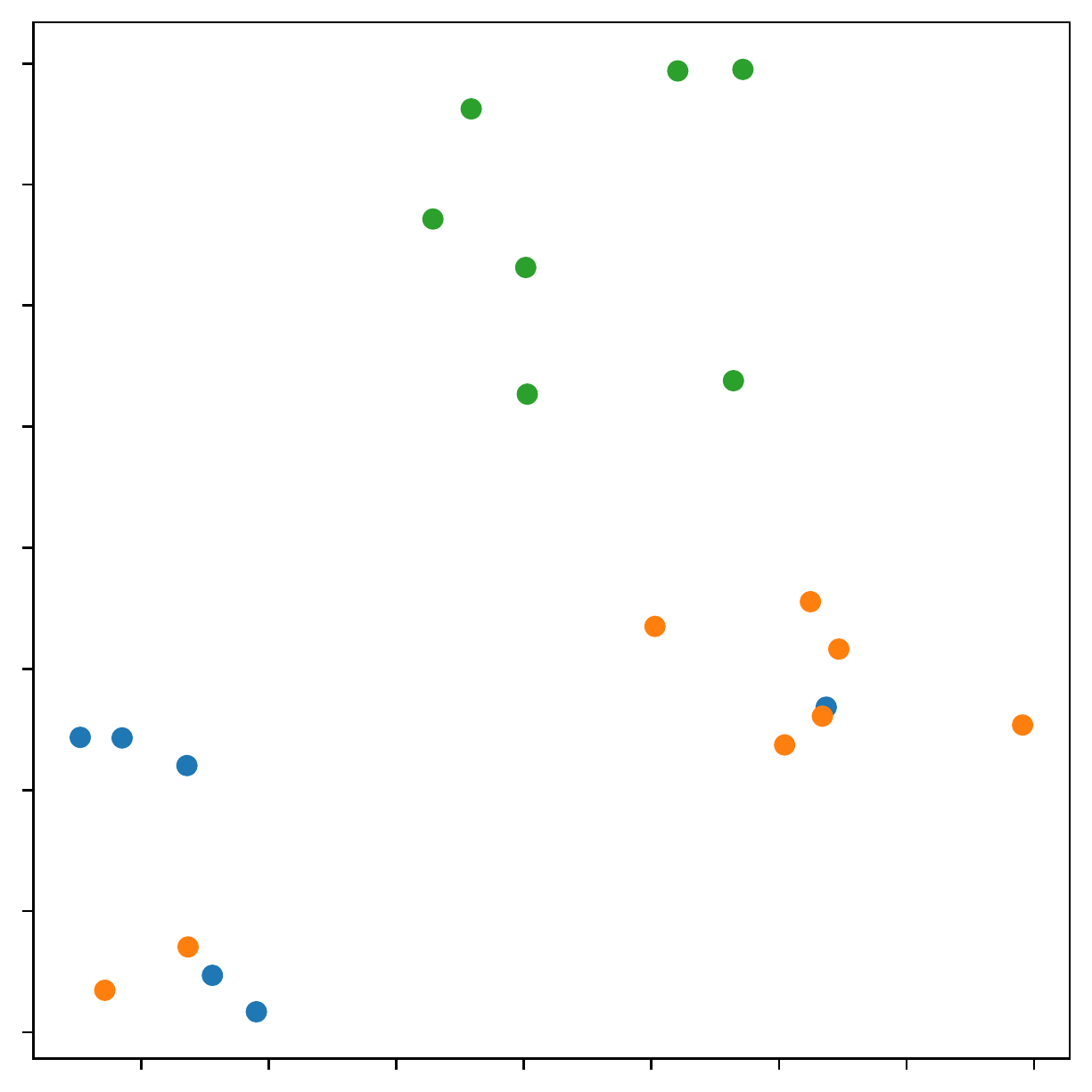}}	  
    &	 \makecell[l]{$\min_{\bm{c},\bm{\theta} }\; \; \;\sum_{k=1}^{K}\sum_{i=1}^{N} l( \bm{x_{i}},y_{i}, \bm{\theta_k})  \; c_{ik}$ \\
    $\text{s.t.} \; \; \;\;\;\; \;\;  \sum_{k=1}^{K} c_{ik} = 1, \;\; i \in \overline{N}$	\\
    $\; \; \;\;\;\;\;\;\;\; \;\; \;\;  c_{ik} = \{ 0 , 1\}, \;\; i \in \overline{N}, \; \;  k \in \overline{K}$       } \\ \midrule
    
    \makecell[c]{Closest Center  \\	  \includegraphics[scale=0.3, valign=t]{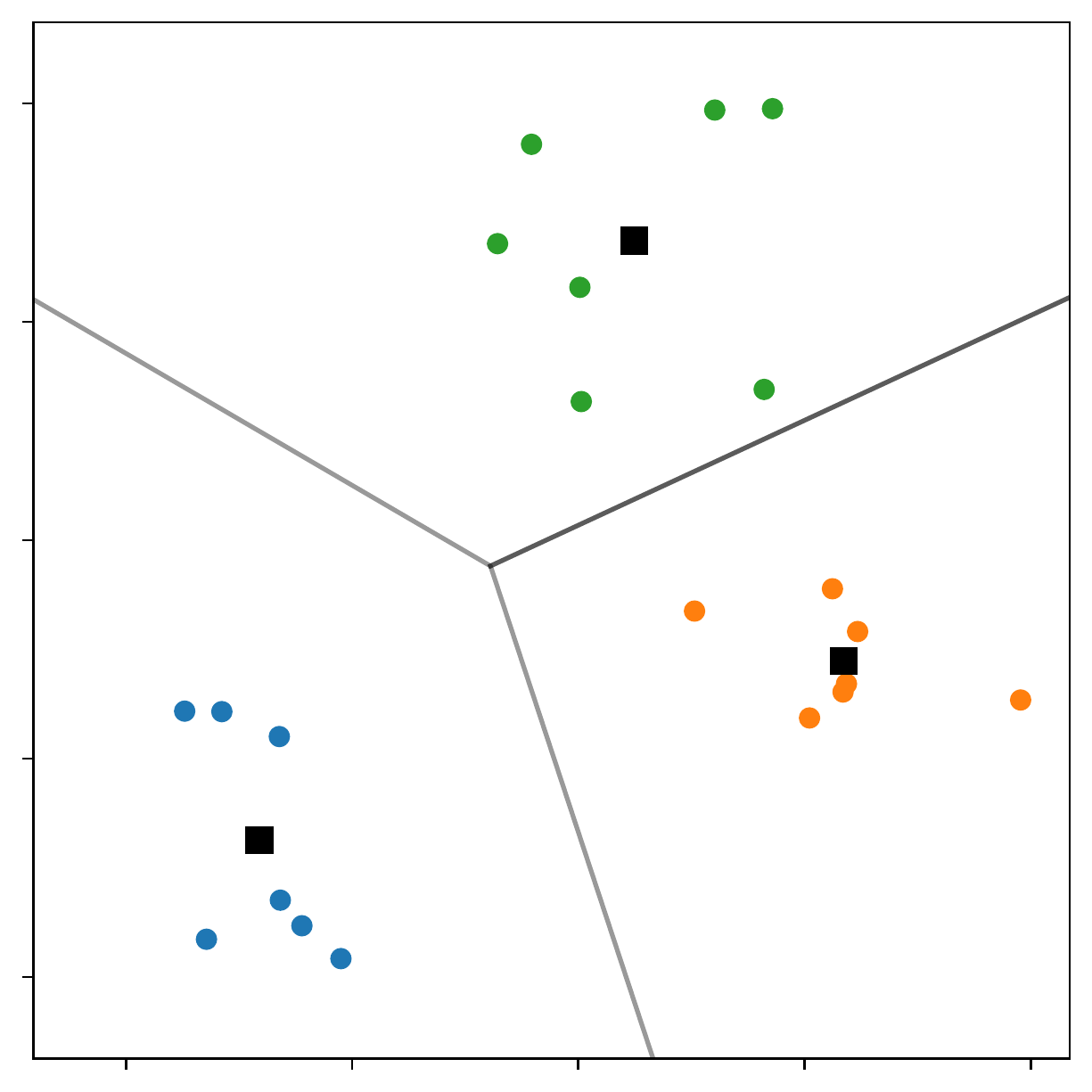}}
    & \makecell[l]{
    $\min_{c,\bm{\theta} }\; \; \;\sum_{k=1}^{K}\sum_{i=1}^{N} (l( \bm{x_{i}},y_{i}, \bm{\theta_k})  \; c_{ik} + \lambda \; d_{i} )$ \\
    $ \text{s.t.} \; \; \;\;\;\; \;\; \;d_{i} \geq \| \bm{x_{i}} - \bm{\beta_{k}} \|_1 - M_{2}(1 - c_{ik}),   $ \\
    $ \hfill i \in \overline{N}, \; \; k \in \overline{K}$ \\
    $\; \; \;\;\;\;\;\;\;\; \;\; \;\; d_{i} \geq  0,  \; \; i \in \overline{N}$ \\
    $\; \; \;\;\;\;\;\;\;\; \;\; \;\; \sum_{k=1}^{K} c_{ik} = 1, \;\; i \in \overline{N} $ \\ 
    $\; \; \;\;\;\;\;\;\;\; \;\; \;\;  c_{ik} = \{ 0 , 1\}, \;\; i \in \overline{N}, \; \;  k \in \overline{K}$       
    } \\ \midrule
    
    \makecell[c]{Bounding Box  \\	  \includegraphics[scale=0.3, valign=t]{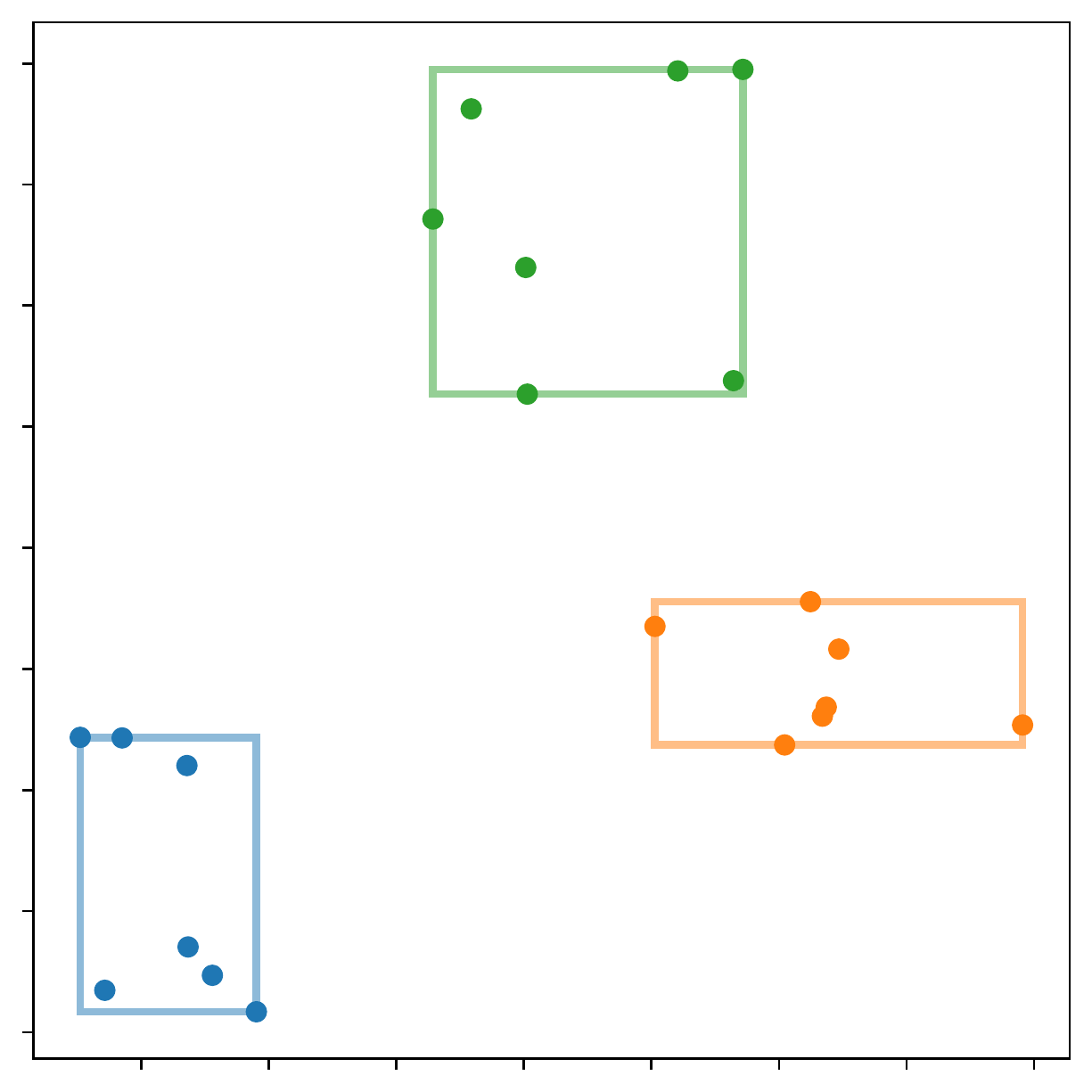}}
    & \makecell[l]{
    $\min_{c,\bm{\theta} }\; \; \;\sum_{k=1}^{K}\sum_{i=1}^{N} l( \bm{x_{i}},y_{i}, \bm{\theta_k})  \; c_{ik}$ \\
    $ \text{s.t.} \; \; \;\;\;\; \;\; \; I_{ikj}=
    \begin{cases}
      1, & \text{if } (x_{kj}^{min} \leq x_{ij} \; \; \bigwedge \; \; x_{ij} \leq x_{kj}^{max}) \ \\
      0, & \text{otherwise}
    \end{cases}, $ \\
    $\hfill j \in \overline{D}, \; \; i \in \overline{N}, \; \; k \in \overline{K}$ \\
    $\; \; \;\;\;\;\;\;\;\; \;\; \;\; c_{ik} = \bigwedge_{j \in J} I_{ikj}, \; \; i \in \overline{N}, \; \; k \in \overline{K}$ \\
    $\; \; \;\;\;\;\;\;\;\; \;\; \;\; x_{kj}^{max} >  x_{kj}^{min},  \;\; j \in \overline{D}, \; \;  k \in \overline{K}$ \\
    $\; \; \;\;\;\;\;\;\;\; \;\; \;\; \sum_{k=1}^{K} c_{ik} = 1, \; \; i \in \overline{N} $ \\ 
    $\; \; \;\;\;\;\;\;\;\; \;\; \;\; x_{kj}^{max} , x_{kj}^{min} \in \mathbb{R}, \;\;\; j \in \overline{D}, \; \; k \in \overline{K}$ \\
    $\; \; \;\;\;\;\;\;\;\; \;\; \;\; c_{ik} = \{ 0 , 1\}, \;\; i \in \overline{N}, \; \;  k \in \overline{K}$} \\ 
    \bottomrule
\end{tabular}
\end{center}
\end{minipage}
\label{tab:milp}
\end{table*}

\subsection{Mixed Integer Optimization}
\label{section:mio}

Having presented the two components of our framework - clustering and prediction objective in the previous subsections, we now show how to marry them together to get the desired model. The available model choices in our design space are shown in Table~\ref{tab:designspace}. We can mix and match the three cluster definitions and the two-loss functions to give an array of models tailored to specific problems.


Our general strategy for optimization is to define a set of constraints for each of the three clustering methods and combine it with the previously described supervised objective functions. We then employ MIO to obtain globally optimal results for our models. The general form of the objective function is given in Equation~\ref{eqn:totalloss}. The objective function in this form is non-linear. Therefore, we reformulated the cost function using the ``big-M" method as follows:





\begin{equation}
\begin{split}
    &\min_{c,\bm{\theta} }\; \; \;\sum_{k=1}^{K}\sum_{i=1}^{N} e_{ik}\\
    &\text{s.t.} \; \; \;\; l( \bm{x_{i}},y_{i}, \bm{\theta_k}) - e_{ik} \leq M*(1 - c_{ik}), \\
    & \;\; i \in \overline{N}, \; \;  k \in \overline{K} \\
    & \; \; \;\;\;\;\;\; e_{ik} \geq 0, \;\;\;  \; i \in \overline{N}, \; \; k \in \overline{K} \\
    & \; \; \;\;\;\;\;\; c_{ik} \in \{ 0 , 1\}, \;\;\;\; i \in \overline{N}, \; \; k \in \overline{K} \\
\end{split}
\label{eqn:cc}
\end{equation}

With this reformulation, we forced the new variable $e_{ik}$ to take the value of $l( \bm{x_{i}},y_{i}, \bm{\theta_k})$ when $c_{ik} = 1$. Additionally, when $c_{ik} = 0$, minimization of the objective function along with the constraints ensured that $e_{ik} = 0$, i.e., when an observation does not belong to a cluster $k$, it does not incur prediction error w.r.t. that cluster. We remark that the choice of the big-M is critical in ensuring the reformulation works as expected. 


In clustering, the main objective is to associate each point to one cluster and further add clustering type-specific restrictions. This is achieved by appropriately placing constraints on the indicator variables $c_{ik}$. We describe these constraints in Table~\ref{tab:milp}.

In the case of (1) arbitrary clustering: we used constraints to make sure that a point is assigned to only one cluster; (2) closest center clustering: we introduced variables $d_{i}$ to capture the distance between a point and the cluster center $\bm{\beta_{k}}$. We added this variable to the objective function to ensure that points are assigned to their closest cluster center. A hyperparameter $\lambda$ was also used to trade-off between the supervised error and point to cluster center distances. In such a formulation, the indicator constraints along with the minimization criteria ensured that when $c_{ik} = 1$ then $d_{i} = \| \bm{x_{i}} - \bm{\beta_{k}} \|_{1}  $. We choose the L1 norm distance metric to compute distances between points and cluster centers to have computationally more feasible linear programming formulation; (3) bounding box clustering: we employed additional variables $x_{kj}^{max}$ and $x_{kj}^{min}$ to define the edges of the bounding box, and indicator variables $I_{ikj}$ to force points that belong within these boundaries to belong to that cluster.

With the appropriate choice of the loss function, which was MAE for regression and L1 regularized SVM loss for classification, we had reformulated our overall problem as a MILP. However, such a MILP-based approach is NP-hard and a very difficult problem to solve~\cite{Lau}. This methodology is only practical with small datasets with a few hundred observations. Therefore, we describe greedy approaches to optimize our models in the following subsection. We used our MILP based solutions to benchmark these greedy methods with synthetic datasets.  

\subsection{Greedy Optimization}



We were inspired by the Majorization-Minimization (MM)~\cite{hunter_lange,mm_tut, gmm} algorithm framework to build our greedy methods. Fundamentally, the MM prescription for constructing algorithms is based on the principle of identifying a suitable, ``easy to optimize" surrogate function to assist in the optimization of a non-convex objective. The algorithms iteratively optimize a sequence of these surrogate functions to drive the optimization of the original objective.

Formally, in a minimization task for an objective function $f(\theta)$ w.r.t. parameter $\theta$, we have a surrogate majorizing function $g_t(\theta)$ at the $t$-th iteration satisfying the following: (1) touching condition $f(\theta^{(t)}) = g_t(\theta^{(t)})$ which ensures that both functions have the same value at $\theta^{(t)}$; and (2) condition that $g_t(\theta)$ majorizes $f(\theta)$, i.e.,  $f(\theta) \leq g_{t}(\theta)$. At each time step, we minimize the majorizing function to obtain the value of parameters for the next time step given by $\theta^{(t+1)}$. This process is repeated to drive the original objective to a minima, but without assured convergence to global optima. The commonly seen expectation-maximization (EM) approach is a special case of the MM algorithm.


We briefly describe our algorithm and the elements of the MM framework that we adopted in our greedy search for clusters with the joint optimization of the supervised loss. We used the `Predictive-clustering' algorithm to wrap the overall procedure and an `assignment' subroutine to assign points to clusters.

\begin{enumerate}
\item \textbf{Predictive-clustering algorithm:} This algorithm, as shown in Algorithm~\ref{alg:main} runs an iterative procedure with two steps until the convergence of loss function up to a threshold. First, the cluster assignment variables are randomly initialized. This is followed by an iterative procedure that involves: (1) optimizing the error function after fixing the point-cluster assignments to learn a new set of parameters $\theta^{(t)}$ (line~\ref{alg:mainline4} in Algorithm~\ref{alg:main}); and (2) reassigning points to clusters based on the new parameters and clustering criteria (line~\ref{alg:mainline6} in Algorithm~\ref{alg:main}).

\algnewcommand{\Initialize}[1]{%
  \State \textbf{Initialize:}  \parbox[t]{.8\linewidth}{\raggedright #1}
}
\algnewcommand{\LineComment}[1]{\Statex \hspace{0.6cm}\(\triangleright\) \textit{#1}}
\begin{algorithm}[t!]
\caption{Predictive Clustering Algorithm}\label{alg:main}
\begin{algorithmic}[1]
\Require $Data: (\bm{x_{i}},y_{i})$
\Initialize{Cluster assignment $ C_{ik}^{(0)}$}
\State $t \gets 0$
\While{Convergence of Loss}
    \State $\bm{\theta_{k}^{(t)}}  \gets $ Optimize$(l( \bm{x_{i}},y_{i}, \bm{\theta_k^{(t-1)}}) , c_{ik}^{(t)})$
    \label{alg:mainline4}
    \LineComment{Optimize() function optimizes the loss function given $c_{ik}^{(t)}$}
    \State $c_{ik}^{(t+1)} \gets$  Assignment-function$( \bm{x_{i}}, \bm{\theta_k^{(t)}}) $
    \label{alg:mainline6}
    \LineComment{Assignment-function returns a new assignment}
    \State $t \gets t+1$
\EndWhile
\State \Return $c_{ik}, \bm{\theta_{k}}$
\end{algorithmic}
\end{algorithm}

\begin{algorithm}[t!]
\caption{Assignment Function }\label{alg:assign}
\begin{algorithmic}[1]
\Require $ (\bm{x_{i}},y_{i}), \bm{\theta_k^{(j)}}$

\State $c_{ik}^{(t+1)} = \{\mathbbm{1}_{k = k_i^*} \; \rvert \; k_i^* = \text{arg} \min_{k} \; l( \bm{x_{i}},y_{i}, \bm{\theta_k^{(t)}}) \} \; \;  \forall i \;  \in \overline{N}$
\label{alg:assign1}
\State $\bm{z_{k^*}} \gets $ Centroid$(\bm{x_{i}}, c_{ik}^{(t+1)})$
\LineComment{Centroid() function computes centroids $(\bm{z_{k^*}})$ based on assignments $c_{ik}^{(t+1)}$}
\If{Closest center clustering}
    \State $c_{ik}^{(t+1)} = \{\mathbbm{1}_{k = k^{*}} \; \vert \; k^{\dagger} = \text{arg} \min_{k^*} \| \bm{x_{i}} - \bm{z_{k^*}}\|_{2} \} \; \forall \; i \in \overline{N} $
\label{alg:assign2}
\LineComment{Assigning points to their closest centroid}
\ElsIf{Bounding box clustering}
    \State $c_{ik}^{(t+1)} = \{\mathbbm{1}_{k = k^{\dagger}} \; \rvert \; k^{\dagger} = \text{arg} \min_{k^*} \| \bm{x_{i}} - \bm{z_{k^*}}\|_1 \} \; \forall \; i \in \overline{N} $
\label{alg:assign3}
\LineComment{Assigning points to have them inside bounding boxes}

\EndIf
\State  \Return $c_{ik}^{(t+1)}$

\end{algorithmic}
\end{algorithm}

We utilized the more traditional MSE loss function for regression and L2-regularized SVM for classification tasks in our greedy methods. As a result, when the cluster assignments were fixed, the overall objective reduced to a per cluster supervised learning (regression or classification) problem with smooth convex loss functions that were easy to solve. Consider the regression case when the cluster assignments are fixed,
\begin{multline}
    g_t(\theta) = \bigg\{ \sum_{k=1}^{K}\sum_{i=1}^{N} (y_{i} - \bm{\theta_{k}'x_{i}})^2 \rvert c_{ik}^{(t)} \\ 
     \;\;, i \in \overline{N}, \; \;  k \in \overline{K}\bigg\}
\label{eqn:MMloss}
\end{multline}

Under the MM framework definitions, the function $g_t(\theta)$ in Equation~\ref{eqn:MMloss} is our easy to solve convex surrogate function. Optimizing this function gives us the best regression weights for the next iteration $\theta_k^{(t)}$. This step is followed by the reassignment step where the `Assignment' function is called to return the indicator variables $c_{ik}^{(t+1)}$ for the next iteration.

\item \textbf{Assignment function: }The reassignment step is more complicated since it needs to address the different cluster assignment criteria. Here, the cluster-specific parameters ($\theta_{k}^{(t)}$) are fixed. This subroutine as shown in Algorithm~\ref{alg:assign} reassigns a point to a different cluster if it has a lower prediction error when assigned to that cluster. Continuing the regression example, the new assignments are as follows:
\begin{equation}
    c_{ik}^{(t+1)} = \{\mathbbm{1}_{k = k_i^*} \vert k_i^* = \text{arg} \min_{k} l( \bm{x_{i}},y_{i}, \bm{\theta_k^{(t)}}) \} 
    \label{eqn:MMassign}
\end{equation}    

When the function stops at this step (line~\ref{alg:assign1} in Algorithm~\ref{alg:assign}) and returns the new assignment variables $c_{ik}^{(t+1)}$, we arrive at the result for the arbitrary clustering case. 
These new assignment variables can now be used to define the surrogate function for the next iteration by plugging them into Equation~\ref{eqn:MMloss}. Precisely, this assignment step as described in Equation~\ref{eqn:MMassign} ensures the ``touching condition'' for the next time step under the MM framework definition. A similar procedure is described in a recent work by Manawami and Sastry~\cite{Manwani} (called the K-plane regression algorithm) to solve the traditional CLR problem, but they do not make this connection between their algorithm and the MM approach.

Furthermore, we extend this assignment subroutine to address the other clustering types by slightly deviating from the MM framework. The function achieves the other clustering methods by either: (1) computing the new cluster centroids (with variables $c_{ik}^{(t+1)}$), and reassigning all points to the closest cluster center using the Euclidean distance metric to get the closest center clustering (line~\ref{alg:assign2} in Algorithm~\ref{alg:assign}); (2) computing the new centroids as above but now assigning points to the closest cluster centers using the L1 norm distance metric to obtain approximate bounding box clustering (line~\ref{alg:assign3} in Algorithm~\ref{alg:assign}). This is because when Voronoi diagrams are plotted with the L1 norm distance, the polygon edges are axis-parallel, giving an approximate bounding box shape. 


\end{enumerate}

\section{Results}

In this section, we experimentally investigate the ability of our models to converge to the ground truth using synthetic datasets. We compare our greedy methods with the MILP-based approach, and report the results from our experiments. We then report results for four real-world datasets to motivate and demonstrate the applicability of our models.  

\subsection{Performance Evaluation}
\label{perfeval}
In this subsection, we benchmark the performance of our greedy methods with the MILP-based approaches and empirically show that these methods perform well using synthetic datasets. We obtain globally optimal solutions with MILP-based approaches, but they are not salable. In contrast, greedy methods may not guarantee global optimality, but they are more practical for real-world data. Therefore, we designed our experiments intending to understand (1) how well these greedy methods learn the underlying true generative model for several synthetic datasets; and (2) how time-efficient they are compared to MILP methods.

\begin{figure*}[t]
\begin{centering}
\subfloat[Regression task]{\includegraphics[width = 7.8cm]{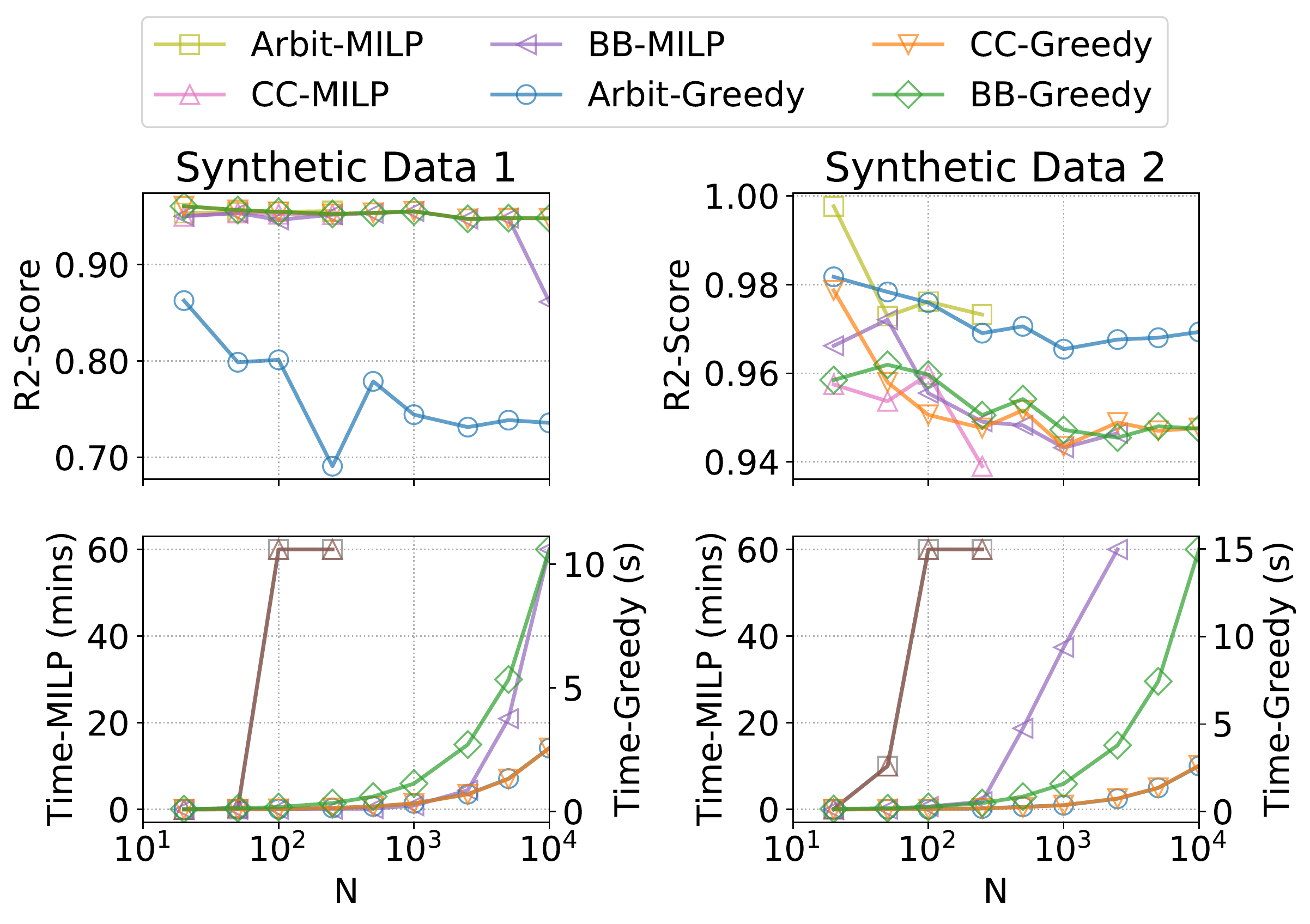}\label{fig:regeval}}  \hspace{0.09cm} 
\subfloat[Classification task]{\includegraphics[width=7.8cm]{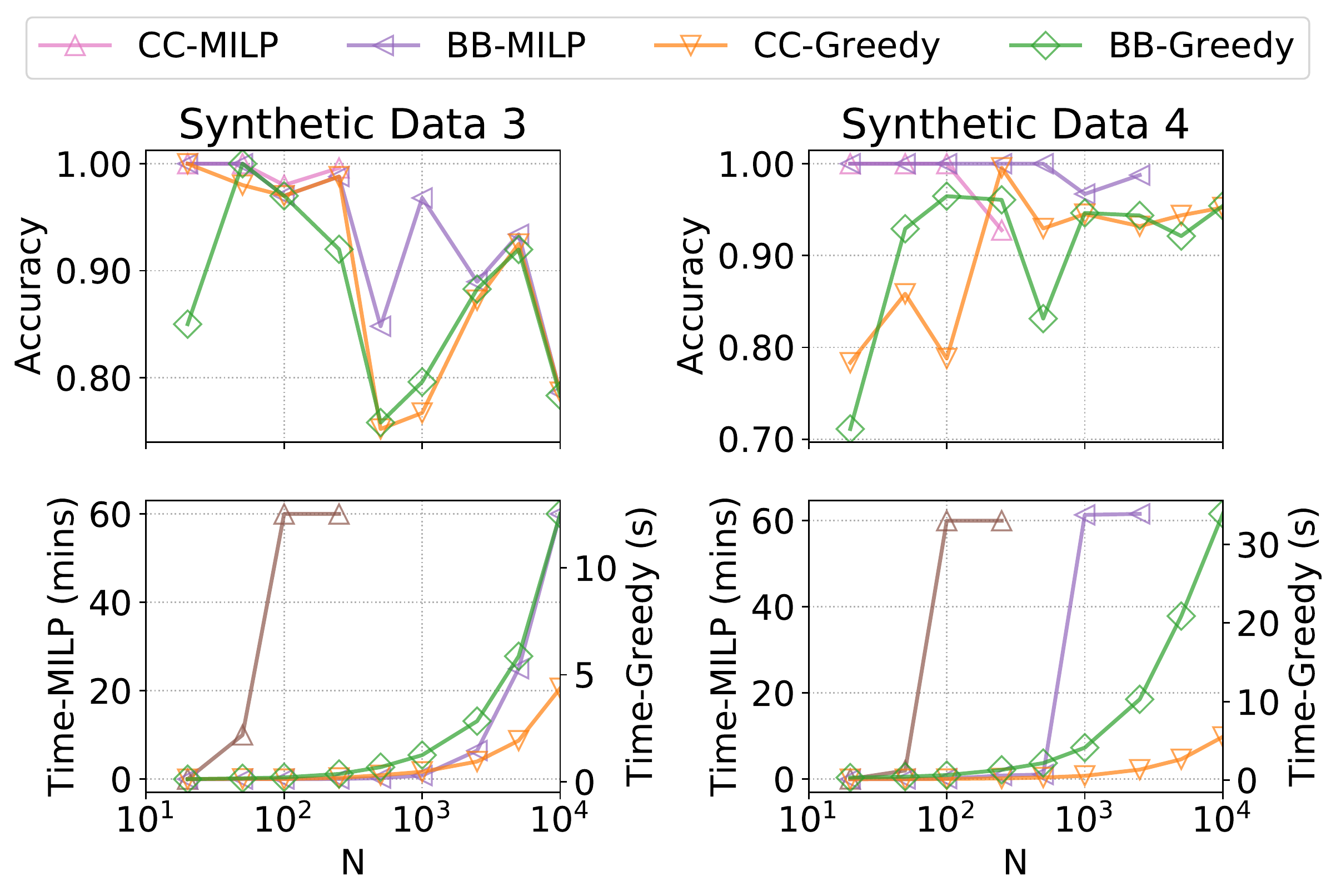}\label{fig:classeval}} \\
\par\end{centering}
\caption{Performance of the MILP-based and greedy algorithms with (a) regression and (b) classification tasks for the different clustering types with 4 synthetic datasets (two per supervised objective). Size of the dataset is varied from $N$ 20 to $10^4$ and the time taken (shown in minutes for MILP and in seconds for greedy approach) to run the models and the resulting $\text{R}^2$ score/Accuracy obtained are reported. 
}
\label{fig:eval}
\end{figure*}




To implement our MILP-based approach, we employed the commercially available Gurobi solver~\cite{gurobi} which is free for academic use. We evaluate all our models in a desktop computer with an 8-core CPU at 3.2 GHz and 8 GB memory. For our MILP models, we fixed the exit optimality-gap threshold at $5\%$. We also prescribed an upper limit of 1 hour running time per experiment. On the other hand, each evaluation was carried out for our greedy methods by averaging the results over ten independent runs of these models on the synthetic data.

Since we aim to understand the ability of our models to learn the underlying ground truth, we chose different generative models to construct the synthetic datasets. First, we took two feature variables and generated reasonably well-separated clusters of points in this feature space with the number of clusters $K \in \{2,3\}$. Then, we used cluster-specific regression weights chosen randomly to give two datasets for the CLR task (Gaussian noise was also added). Similarly, two datasets for the CLC task were generated with a binary-classification objective per cluster (hyperplanes separating classes are different for different cluster) with some noise (class labels assigned randomly). Finally, we ran our experiments by varying the size of the data $N$ from 20 to $10^4$. 


{\bf Regression case:} We report the results in Figure~\ref{fig:regeval} with the experimental setup described above for the two datasets. Here, we compared the MILP-based and greedy methods for each of the three clustering types. The evaluation metric we utilized was the overall $R^2$ score to measure the goodness of regression fit across the different clusters.

We observed that MILP for the closest center and arbitrary clustering methods were not feasible for more than $N = 250$ observations. In fact, the optimality gap was over $30\%$ at the 1-hour exit condition for the $N=250$ case, and hence, we do not report results for larger $N$ values. Interestingly, we found that MILP for the bounding box method is much more scalable. Moreover, it is evident that the greedy methods perform well in most cases and are comparable to the MILP methods. We also note that the performance of greedy CLR with arbitrary assignment is good for the synthetic data 2 but does poorly for the other data. This is because overlapping regression planes, although not representative of the underlying trends in the data, can sometimes lead to a lower error due to the added noise.



{\bf Classification case:} Similarly, we report the classification results in the Figure~\ref{fig:classeval} for our four models - closest centroid and bounding box clustering with greedy and MILP approaches. Here, we used the accuracy metric to evaluate our models.  It is evident that the greedy methods perform as well as the MILP on most occasions. While it may be concerning that the greedy methods performed poorly when $N \leq 50$, we remark that our greedy methods can sometimes reach a solution with all points being in the same cluster when the number of observations are very small, resulting in a poor local minima. \\ 

In conclusion, these evaluations show that the greedy algorithms provide scalable solutions that are a good approximation to the MILP-based methods. Furthermore, with the bounding boxes MILP method succeeding to attain the 5\% optimality gap threshold even in cases with more than 1000 observations in the data, we found that it is significantly more scalable when compared with the MILP for other clustering methods. We believe that defining clusters as bounding boxes introduces much stronger constraints (or cuts on the feasible space), resulting in much-reduced search space for the MILP solvers.

\subsection{Case Study}
In this section, we illustrate the relevance of the array of models available in our design space by analyzing four different real-world datasets picked from a diverse set of domains. Each of these application problems asks a very different question, and we show how we can mix and match tools available in our framework to address them. Through these case studies, we aim at exploring our models' ability to (1) scale for large datasets, (2) perform better than the baseline linear models, and (3) provide highly interpretable results that help in uncovering the different underlying modes of behaviors in data. We focus on benchmarking model performance with the Boston housing dataset, interpretability of results with San Francisco crime rate and FAA Wildlife-strike dataset, and the model's ability to scale with the MovieLens 100k dataset.

As a general preprocessing step, we partitioned the data into the train (65\%), validation (15\%), and test (20\%) sets to tune our hyperparameters (with a focus on finding the best $K$ clusters) and report the 5-fold cross-validation results. We used $\text{R}^2$ score and accuracy metric to evaluate our regression and classification models, respectively. Furthermore, we compared the results from our greedy algorithms with baseline Lasso regression and SVM one-vs-all models from the sklearn package in python~\cite{sklearn}. 


\subsubsection{Boston Housing data} 
\label{sec:boston}
We use the popular and well-studied Boston housing dataset to perform a clusterwise regression analysis and benchmark our model’s performance. As mentioned previously, we expect property values to have multiple trends in different parts of the city. This makes Boston housing\footnote{Stat-CMU StatLib Datasets Archive. Boston house-price data. Retrieved from \url{http://lib.stat.cmu.edu/datasets/boston}} an interesting and relevant dataset to analyze using CLR and understand if our models can identify various trends.

\begin{table*}[t]
\caption{Cluster-specific feature averages (ft) and regression weights (wt) for the Boston housing data analysis}   
\begin{minipage}{\textwidth}
\begin{center}
\setlength\tabcolsep{5pt}
\begin{tabular}{lrrrrrrrrrrrr}
\toprule
 &  \multicolumn{2}{r}{Cluster 1}  &  \multicolumn{2}{r}{Cluster 2}  &    \multicolumn{2}{r}{Cluster 3}  &    \multicolumn{2}{r}{Cluster 4}  &   \multicolumn{2}{r}{Cluster 5}  & \multicolumn{2}{r}{Cluster 6}  \\ \cmidrule{2-13} 
Feature &     ft &   wt &     ft &   wt &     ft &   wt &     ft &   wt &     ft &   wt &     ft &   wt \\
\midrule
   CRIM &   0.5 &      -1.56 &   0.6 &       2.17 &   3.2 &       5.78 &   0.2 &       2.24 &  10.5 &      -0.69 &   0.1 &       1.61 \\
     ZN &   6.6 &      -2.72 &   0.3 &       4.64 &   0.0 &       0.00 &   0.8 &      -3.68 &   0.0 &       0.00 &  37.9 &       0.49 \\
  INDUS &   7.1 &       4.53 &  12.4 &       1.71 &  19.9 &      -0.88 &   8.6 &      -1.97 &  18.5 &      -2.50 &   4.7 &      -1.52 \\
   CHAS &   0.2 &       0.11 &   0.1 &       0.59 &   0.2 &       1.81 &   0.1 &       0.49 &   0.0 &      -0.40 &   0.0 &       0.68 \\
    NOX &   0.5 &      -3.75 &   0.5 &      -5.88 &   0.6 &      -4.23 &   0.5 &      -1.43 &   0.7 &      -3.34 &   0.4 &      -1.98 \\
     RM &   7.1 &       5.98 &   5.9 &       0.91 &   6.3 &      -4.37 &   6.1 &       4.42 &   5.9 &      -1.51 &   6.6 &       7.13 \\
    AGE &  76.8 &      -2.99 &  90.4 &      -1.89 &  83.5 &       0.68 &  63.1 &      -2.06 &  92.0 &       1.64 &  32.2 &      -1.25 \\
    DIS &   3.0 &      -5.64 &   3.2 &      -1.28 &   2.5 &     -11.48 &   3.7 &      -3.33 &   1.9 &      -1.31 &   6.3 &      -1.36 \\
    RAD &   4.7 &       6.64 &   4.4 &       0.85 &  16.6 &       2.17 &   4.5 &       3.02 &  21.0 &       1.99 &   4.2 &       1.72 \\
    TAX & 289.1 &     -11.61 & 330.9 &       0.42 & 600.2 &       0.06 & 311.9 &      -2.16 & 631.8 &      -2.26 & 295.4 &      -2.59 \\
PTRATIO &  16.3 &      -4.66 &  19.2 &      -1.24 &  20.2 &       5.49 &  18.7 &      -1.30 &  19.5 &      -2.78 &  17.4 &      -0.46 \\
      B & 383.9 &      -0.43 & 367.4 &       0.42 & 386.5 &      -2.26 & 391.6 &      -1.36 & 273.7 &       0.42 & 389.5 &       5.44 \\
  LSTAT &   7.0 &      -7.89 &  17.3 &      -2.44 &  12.6 &      -8.85 &  11.7 &       1.07 &  20.6 &      -4.43 &   7.1 &       0.24 \\ \midrule
  MEDV (y)&  33.1 &      &  18.3 &       &  22.2 &      &  21.4 &      &  15.0 &       &  27.3 &       \\
\bottomrule
\end{tabular}
\label{tab:boston}
\end{center}
\end{minipage}
\label{tab:bostonHousing}
\end{table*}

The dataset is small and has $N=506$ observations with 13 features. The variable for prediction is the median value of houses per census tract. The list of features, along with the prediction variable, is shown in Table~\ref{tab:boston}. A description of these features is standardly found across articles~\cite{opc}. We used our greedy methodology for the CLR-CC model to train this dataset. Our choice was based on the idea that the cluster centroids can help understand the average socio-economic and structural feature values in a cluster to explain the cluster-specific regression trends.   

We trained our model with $K \in \{2,...,7\}$ clusters. Best results were found with 6 clusters with an out-of-sample $\text{R}^2$ score of 0.8622. This is very close in comparison with the average test $\text{R}^2$ score of 0.863 reported by by authors in ~\cite{opc} with their optimal OPC model for the same dataset with 6 clusters.

We report the feature averages and regression weights for each cluster in Table~\ref{tab:boston}. It is evident from cluster 5 that the property values are lowest for old houses in areas with a very high crime rate. We generally observe that the prices inverse with a decrease in crime (cluster 5, for example). Yet, in cluster 3, both property values and crime are high, and the variables are positively related (weight of 5.78), which is against the usual trend. This could potentially be a cluster of housing properties in Downtown where crime is generally high, and houses tend to be very expensive besides being old and small (also observed in Table~\ref{tab:boston}). Another pattern worth noting is in cluster 1, which has the highest average housing price. Here, the costs increase steeply when the number of rooms increases and the educational environments improve (indicated by the PTRATIO variable). Overall, our model is able to pick up on the different modes in the data and discover unique patterns.

\subsubsection{FAA Wildlife-Strike data}

\begin{figure*}[t]
    \centering
    \includegraphics[scale=0.34]{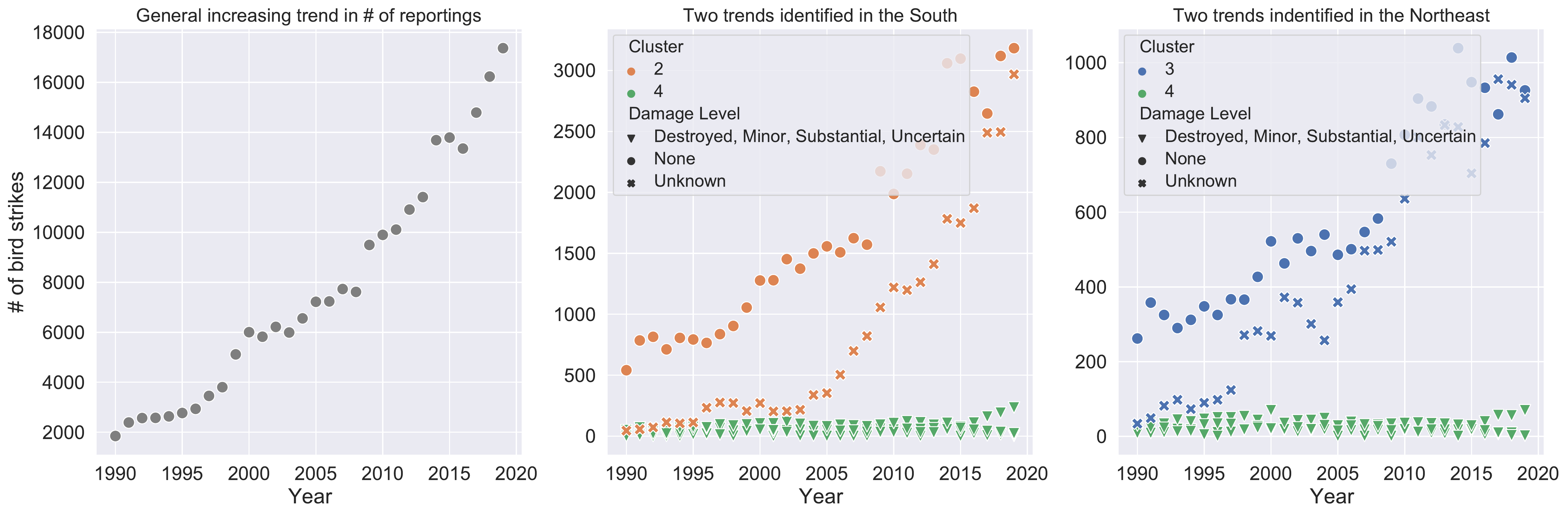}
    \caption{General linear trend observed (left) versus multiple trend identified using our regression-BB clustering model in the South (center) and Northeast regions (right) in the US. 
    For the South region (center), points denoted in orange (cluster 2) have a slope of 746.6 w.r.t to the year feature (increasing trend in reporting when the damage level is none and unknown) whereas the ones denoted in green (cluster 4) have a slope of 2.3 (i.e. the curve is flat where damage levels are destroyed, substantial, minor, or uncertain.}
    \label{fig:faaplot}
\end{figure*}

The FAA Wildlife-strike database contains\footnote{Federal Aviation Administration. FAA Wildlife Strike Database. Retrieved from \url{https://wildlife.faa.gov/home}} the records of all reported aircraft-wildlife strikes (mostly bird strikes) in the US in the last three decades. A general upward trend in the number of bird strikes has been observed over the years, as shown in Figure~\ref{fig:faaplot}. This could be caused by many factors like increased flights and/or birds, or increased reporting every year. We were motivated to explore this dataset with predictive clustering to answer some of these questions.

We were mainly interested in two sets of feature variables: `level of damage' caused to the aircraft due to the bird strike and the `region' in the US where it took place. In the database, we found six levels of damage: minor, substantial, uncertain level, destroyed, unknown (damage not reported) and none (or no damage); and five regions in the US: Midwest, Northeast, South, West, and unknown (when the region is not known). These indicator levels were encoded as binary variables to generate our features. For example, variable `South' = 1 when a bird strike took place in the South region of the US, and 0 otherwise. Finally, after the preprocessing and transformation steps, we grouped the individual bird strike records per year of the strike, damage levels, and the US regions. The resulting count of records, an aggregate of the number of strikes w.r.t to the features, was used as the prediction variable.

\begin{figure}[t]
    \centering
    \includegraphics[scale=0.35]{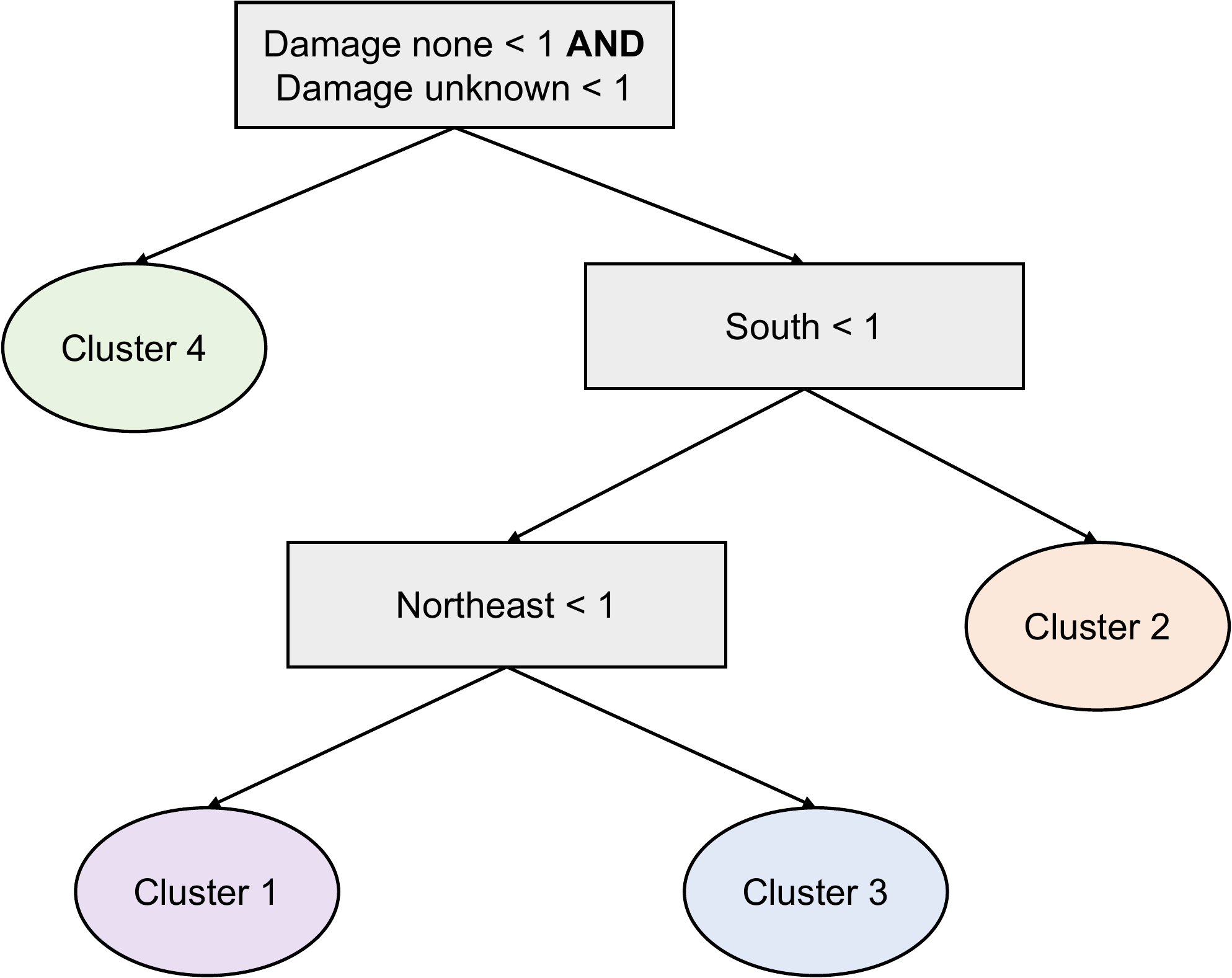}
    \caption{Decision rules based tree architecture representing the 4-clusters obtained in the FAA wildlife-strike dataset analysis}
    \label{fig:faatree}
\end{figure} 



\begin{table*}[t]
\caption{Cluster-specific regression weights for the FAA wildlife-strike data analysis}   
\begin{minipage}{\textwidth}
\begin{center}
\begin{tabular}{lrrrr}
\toprule
              Features &    Cluster 1 &     Cluster 2 &     Cluster 3 &   Cluster 4 \\
\midrule
                  Year &        354.3 &         746.6 &         256.4 &         2.3 \\
             Destroyed &          0.0 &           0.0 &           0.0 &        -8.7 \\
          Minor damage &          0.0 &           0.0 &           0.0 &         9.2 \\
           Damage none &         66.8 &         151.2 &          32.2 &         0.0 \\
    Substantial damage &          0.0 &           0.0 &           0.0 &        -2.6 \\
Damage uncertain level &          0.0 &           0.0 &           0.0 &        -0.9 \\
        Damage unknown &        -67.3 &        -152.2 &         -32.4 &         0.0 \\
               Midwest &         17.3 &           0.0 &           0.0 &        -1.9 \\
             Northeast &          0.0 &           0.0 &           0.0 &        -5.0 \\
                 South &          0.0 &           0.0 &           0.0 &         6.9 \\
        Region unknown &        -77.9 &           0.0 &           0.0 &        -0.8 \\
                  West &         60.4 &           0.0 &           0.0 &         0.7 \\ \midrule
\# of bird-strikes (min-max)  & 1.0 - 2354.0 & 45.0 - 3183.0 & 34.0 - 1039.0 & 1.0 - 235.0 \\
\bottomrule
\end{tabular}
\end{center}
\end{minipage}
\label{tab:faaWeights}
\end{table*}


Our strategy was to use our greedy CLR-BB model to identify highly interpretable clusters and capture different regression lines describing the prediction variable. We have $N = 803$ observations with 12 features in our generated dataset. We trained our model with $K \in \{2,3,4,5\}$ clusters and found that the $K = 4$ case gave the best results in terms of interpretability and performance. The average out-of-sample $\text{R}^2$ score for our model was found to be 0.929, much higher in comparison with the $\text{R}^2$ score of 0.613 for the baseline lasso regression model. We report the regression weights and min-max (`range') of the prediction variable in Table~\ref{tab:faaWeights}. We also leverage our bounding box clustering model's ability to define clusters as a set of decision rules to build a tree-shaped architecture as shown in Figure~\ref{fig:faatree}.

It is evident from Table~\ref{tab:faaWeights} that cluster 4 has a substantially lower number of bird-strikes reported, and the slope w.r.t. to year feature is very small. From the tree in Figure~\ref{fig:faatree}, we realize that this cluster corresponds to the cases when the flight had minor to substantial damage, i.e., when `Damage None' and `Damage unknown' variables are both 0. For the ``no’’ damage case, the model had partitioned the data based on the regions to give 3 clusters. Clusters 2 and 3 correspond to bird strike reporting from the South and Northeast regions. Clearly, the highest reporting was from the South region along with a steeper slope w.r.t. to the year variable. This is reflected in Figure~\ref{fig:faaplot}. The plot for the South region in this Figure~\ref{fig:faaplot}, shows two trends, one which increases steeply corresponding to the ``no'' damage case (cluster 4) while the other that remains flat corresponding to the ``some'' damage case (cluster 2). Similarly, the plot for the Northeast region represents clusters 3 and 4.

Overall, our model was also able to obtain very clear partitions along the features to identify regions of high and low bird strike activity, along with giving clarity on the damage levels in these regions. Because we note that only the curve for the ``no'' damage case is increasing, there is reason to believe that it is only the increased awareness among pilots that has resulted in higher reporting.  




\subsubsection{SF Crime dataset}


Through this case study, our goal was to categorize crime rates in the census tracts in San Francisco into three classes -  high, medium, or low; and understand the relationship between crime in neighborhoods and census features. Moreover, we were motivated to leverage our models to conduct demographic analysis with a geospatial dataset and seek readily interpretable results. We used our greedy classification-bounding box clustering model to conduct this study. Since crime patterns primarily depend on location and intrinsic socio-economic features, we used the bounding boxes method to obtain spatially coherent clusters.
To facilitate this study, we used the Longitudinal Tract Database (LTDB)~\cite{ltdb} to get socio-economic and demographic features for the census tracts (2010) in San Francisco (SF). We then obtained the crime incidents reports from the police department database from 2003 to 2018 from DataSF\footnote{City and County of San Francisco. Police Department Incident Reports. Retrieved from \url{https://data.sfgov.org/Public-Safety/Police-Department-Incident-Reports-Historical-2003/tmnf-yvry}} (an open data portal for SF). We performed several preprocessing steps to prepare the data. First, we computed the per tract count of property and violent crime incidents reported, and then assigned class labels low, medium, and high (corresponding to classes 1,2, and 3) based on it. Next, we used mutual information~\cite{mutual_info} to pick the following census features~\cite{ltdb}: housing units in multi-unit structures (multi), persons in poverty (npov), median house value (mhmval), people with at least a four-year college degree (col), professional employees (prof), per-capita income (incpc), latitude, and longitude (corresponding to the central point in a tract).  


\begin{figure*}[t]
\begin{centering}
\subfloat[Crime rates represented by the three class labels]{\includegraphics[width=7.7cm]{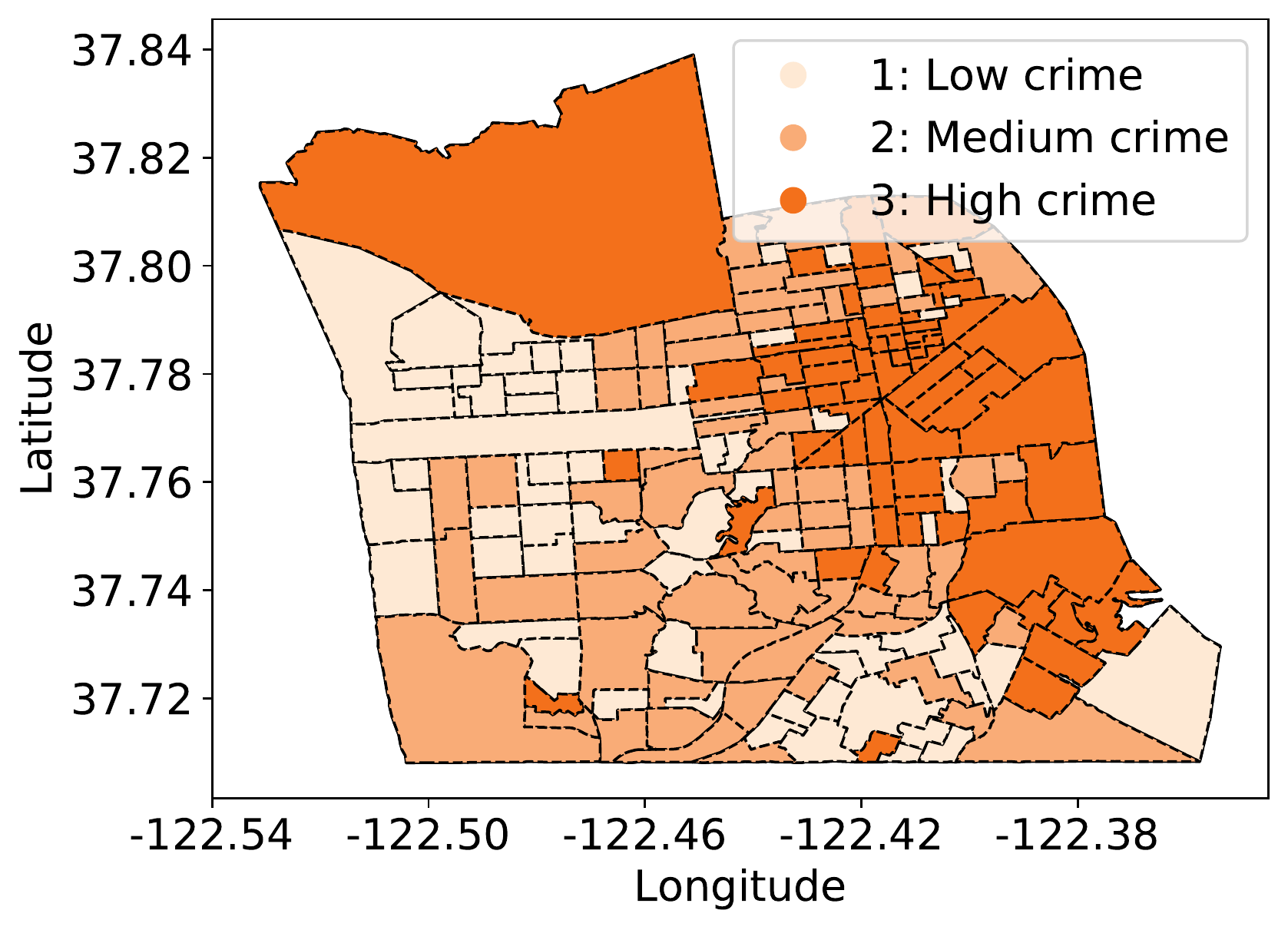}\label{fig:crimeSF}} \hspace{0.09cm} 
\subfloat[Separation of clusters along the tracts]{\includegraphics[width = 7.7cm]{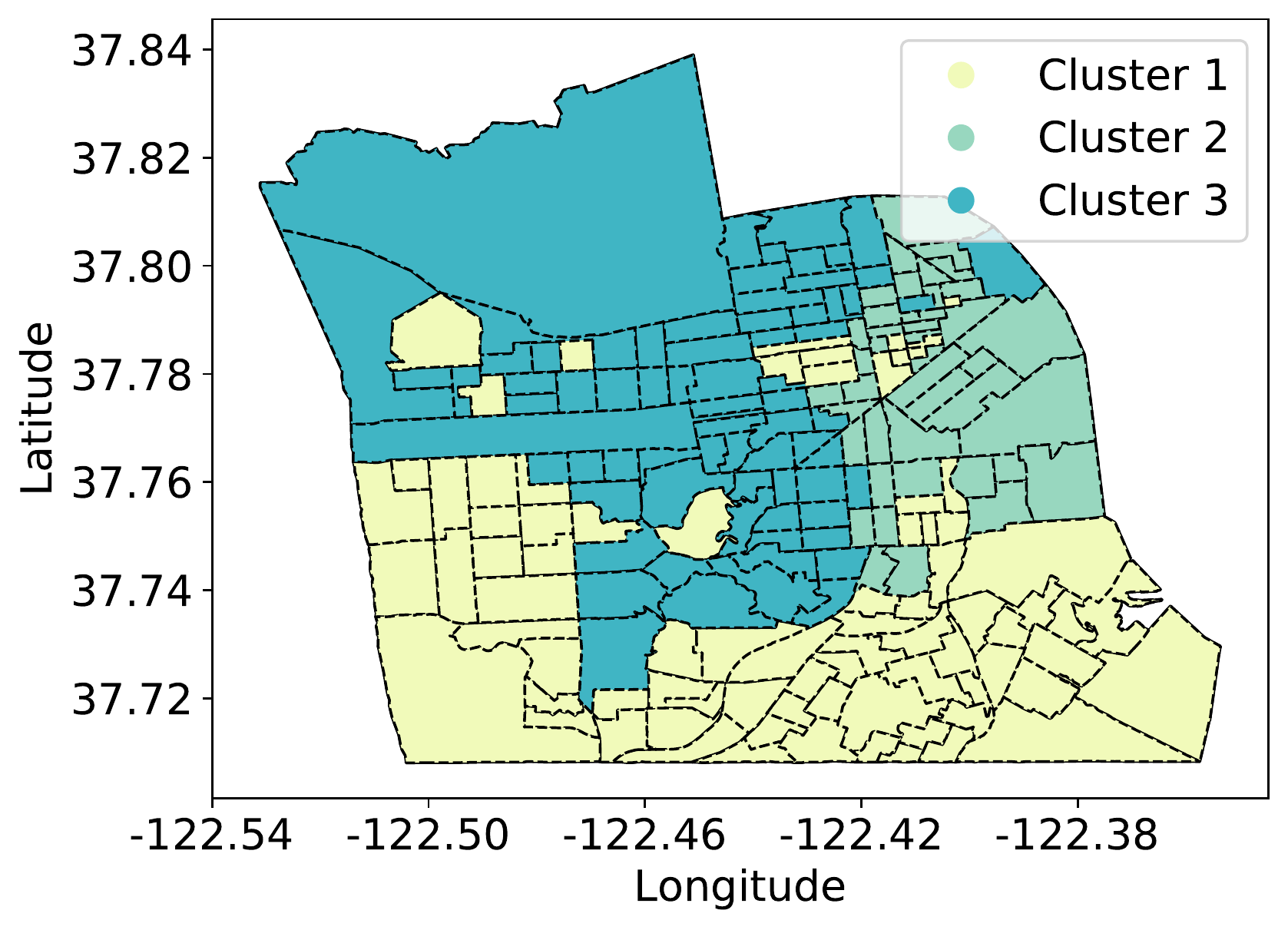}\label{fig:clustersSF}} \\
\subfloat[Boundaries of the 3-clusters across the various feature variables. The lines denote the range of values along each feature and the points in gray represent the feature averages.]{\includegraphics[width=14cm]{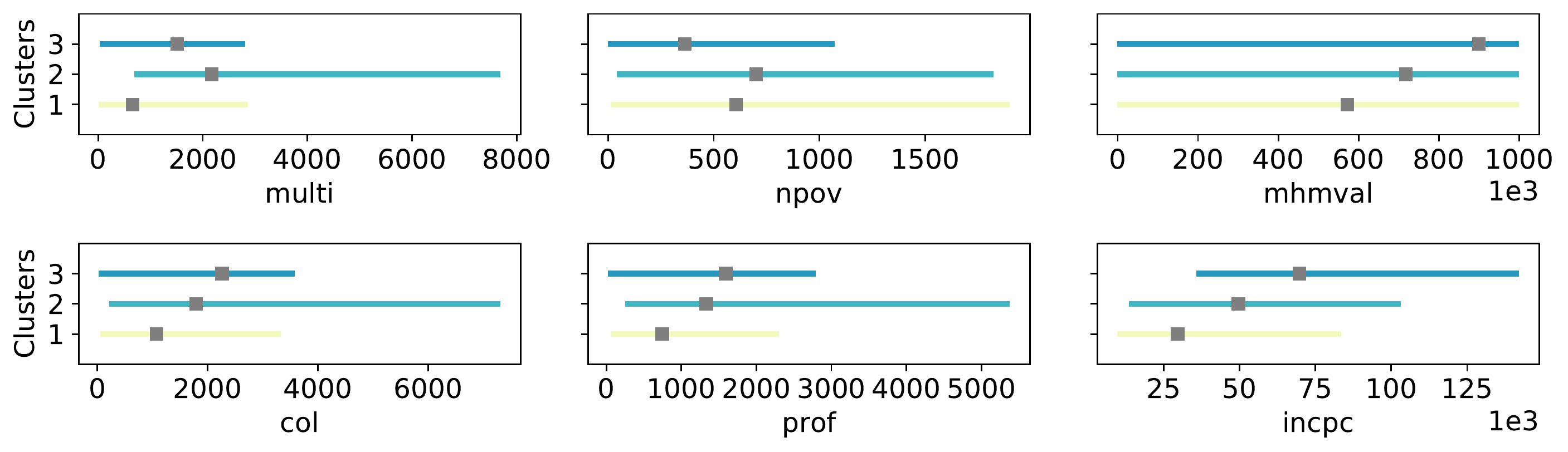}\label{fig:SFfeatures}} 
\par\end{centering}
\caption{The distribution of crime in SF census tracts and the representation of the three spatially coherent clusters obtained for the SF crime dataset analysis
}
\label{fig:crime_map}
\end{figure*}


The generated data is relatively small with $N = 195$ observations with 8 features. We trained the data with our greedy CLC-BB model with $K \in \{2,3,4\}$ clusters and found that the best results were obtained with $K=3$ clusters. The average out-of-sample accuracy was 0.692, an improvement over the baseline linear classification result of 0.558. We report the feature averages and their boundaries for the three clusters in Figure~\ref{fig:SFfeatures}. The prediction label average was 1.86, 2.67, and 1.89 for the three clusters. Clearly, a larger class label average represents a high crime cluster. From Figure~\ref{fig:crime_map}, it is evident that we have two low crime rate clusters (1 and 3).
Moreover, we note that clusters 2 and 3 are similar in being high-income, well-educated, and high property value clusters. However, these clusters have two contrasting modes concerning crime rates. Any traditional unsupervised clustering model like K-means would have put both these groups in the same cluster. Furthermore, a single linear classification model may not efficiently capture such intricate details and multiple modes present in the data.   

With Figure~\ref{fig:crime_map}, we show how we leveraged our bounding box clustering approach to understand the results further. For instance, by connecting census tracts in the map containing the clusters from Figure~\ref{fig:clustersSF} with that of the crime labels per tract shown in Figure~\ref{fig:crimeSF}, we recognize that cluster 2 corresponds to the high crime regions in the Downtown SF. Inner-city areas are expected to have higher crime reporting and a more significant proportion of the educated, high-income working-class population.  

Overall, our clustering method was able to identify spatially coherent clusters while simultaneously recognizing the different modes of crimes observed in these regions. Also, it gave better performance than the baseline, along with interpretable and visually appealing results.

\subsubsection{Movielens data}

Our objective was to use the MovieLens-100K dataset~\cite{ml100k} to perform a recommendation task using user and item features – content-based filtering. We transformed the data to a classification problem and applied our classification-closest centroid greedy model. Although we understand that the state-of-the-art recommendation systems use collaborative filtering or hybrid methods~\cite{CF_survey,rec}, we use our methodology as a proof of concept to drive that predictive clustering models can be used as a first step in exploratory data analysis. Moreover, by using a large dataset for this case study, we could also complement the other three analyses that used relatively smaller datasets.

The MovieLens dataset contains information of 943 users, rating a fraction of the 1682 movies list available. To prepare our data, we went through several pre-processing steps. First, we identified the top 10 genres from a list of 19 genres that cover more than 85\% of the movies in the list. We used these genre indicator variables as our movie features. Additionally, we obtained movie information like popularity indicator, number of votes, vote average, and revenue-budget ratio from the IMDB database to supplement our movie features. Second, we used the user's gender, and age (after binning the age) features for user description. Finally, we merged the two datasets to obtain our user-item rating dataset. Since the ratings are from 1-5, we threshold the ratings at 4, i.e., a rating greater than or equal to 4 got assigned to class 1 (recommend a movie), and class 0 otherwise. This generated dataset has more than 85K observations with 21 features.

\begin{table*}[t]
\caption{Summary of the clusters based on the feature averages}   
\begin{minipage}{\textwidth}
\begin{center}
\begin{tabular}{l|lll}
\toprule
Features	&	Cluster 1	&	Cluster 2	&	Cluster 3	\\ \midrule
Gender	&	Mostly women	&	Male	&	Both male and female	\\
Age (years)	&	All age groups	&	20-30 and 45+	&	Less than 20,  30-45	\\
Genre	&	Drama, romance	&	Drama, thriller, action	&	Comedy, horror	\\ \midrule
y (label) & 0.57 & 0.63 & 0.41 \\
\bottomrule
\end{tabular}
\end{center}
\end{minipage}
\label{tab:movielens}
\end{table*}

We used our greedy CLC-CC model for this dataset. We tuned the value of clusters $K \in \{2,3,4,5,6,7\}$, and found the best results with $K = 3$. We observed an average test accuracy of 0.652 and test root mean square error (rmse) of 0.589.This is a small improvement compared to baseline classification, which gave an average accuracy of 0.637 and rmse of 0.602. In Table~\ref{tab:movielens}, we report a summarized version of what each cluster represents based on the feature averages found in them. The label’s average indicates how users react to movies that fall in a cluster. For instance, cluster 2 indicates that males between 20-30 and 45+ prefer drama, action, and thriller genres. In addition to this, we found the following variables to be significant based on the weights of SVM hyperplanes found in each cluster: vote average, vote count and age categories in cluster 1; vote count, action and crime genre features in cluster 2; and revenue-budget ratio, romance and thriller genres in cluster 3.  Overall, our model was able to provide interpretable results that helped identify these interesting populations sections or target groups, as evident from Table~\ref{tab:movielens}.


\section{Conclusion}

In this section, we briefly summarize our key contributions and results, and discuss important limitations of our work along with possible future research directions to address them.

\subsection{Summary}

In this article, we began with the observation that clustering for data science has often been viewed through the lens of K-means
and the application of clustering for supervised tasks has largely been overlooked and unexplored. We took a broader outlook towards clustering and introduced a novel generalized framework for predictive clustering to address this deficiency.

In our framework, we presented different perspectives to define clusters and a general approach to combine clustering with various supervised objectives. As a result, an array of models falls out of this framework. Some of these have been previously explored in the literature; however, they were restrictive in their approach and largely application-driven. Furthermore, we presented two methodologies to optimize all models in our framework. Using MILP-based formulations, we ensured global optimization for our models and provided reproducible results. Our highly scalable and relatively efficient greedy algorithms inspired by the Majorization-minimization framework give a good approximation of the benchmark optimal MILP-based solution in instances where comparison was possible.

We also demonstrated the relevance of a predictive clustering framework by analyzing and obtaining results for four unique datasets from a diverse set of domains. Through our case studies, we were able to show how these models were able to detect the different generative modes present in the data and how we can interpret these results. Consequently, we obtained significantly better results compared to baseline regression and classification models.

More fundamentally, we had focused on defining a framework that can break the notion of clustering as an unsupervised learning tool, uncover multiple ``behavioral" modes present in the data, and discover ``hidden" patterns in the supervised sense. As a step towards this direction, we developed a small toolkit of supervised clustering methods, which can potentially be expanded to include many more cluster definitions and supervised objectives. This framework could not only be used as a novel conceptual approach to solve many problems in data science but also outperform traditional linear models to give better results. Moreover, we believe that data scientists and policymakers could efficiently leverage this toolkit to obtain workable solutions that are highly interpretable and can help design policy interventions.



\subsection{Future work}
Overall, we believe that this work brings an alternative broader outlook towards clustering and thereby can act as a catalyst to inspire a range of fascinating extensions and future applications. We list some exciting areas of future work below:
\begin{enumerate}
    \item Expansion of the design space along both dimensions: By defining a generalized framework for optimization for predictive clustering, we enable the scope for expanding the design space along both the clustering type and supervised objective dimensions. For example, unsupervised clustering methods like DBSCAN and spectral clustering can be added to the array of the cluster definitions already a part of the framework. Density-based clustering like DBSCAN could add the flavor of having arbitrary-shaped clusters, unlike the hyper-cuboids of the bounding box and spherical clusters of closest center clustering. Furthermore, several other loss functions can be incorporated into the framework, including 0-1 loss and Huber loss for classification with MILP-based and greedy optimization and cross-entropy loss function with greedy optimization. This would provide a broader toolkit of methods to choose from to tackle the constantly evolving needs in the data science field. More importantly, such a framework will then partially enjoy the non-linearity advantage of random forests and neural networks while still retaining its quality of being highly interpretable.
     
    \item Scalable optimization: As seen previously, MILP-based methods for predictive clustering were not practical to solve in real-time for large datasets but nonetheless provided global optimization. Although we observed some improvement with the bounding box clustering in this aspect, there is undoubtedly a need to address scalability for these models. We believe that further research can utilize decomposition methods, tighter and symmetry breaking constraints, constraint and column generation techniques to strengthen the optimization. This would enable us to exploit the global optimization advantage of mixed integer optimization while being able to scale for large datasets that are generally encountered in the real world.

\end{enumerate}

\bibliographystyle{sn-basic}
\bibliography{sn-bibliography}


\end{document}